\documentclass{article} 
\usepackage{iclr2027_conference,times}

\usepackage{amsmath,amsfonts,bm}

\def\Figref#1{Figure~\ref{#1}}

\def\Secref#1{Section~\ref{#1}}

\def\eqref#1{equation~\ref{#1}}
\def\Eqref#1{Equation~\ref{#1}}

\def\Algref#1{Algorithm~\ref{#1}}

\def\1{\bm{1}}

\def\vzero{{\bm{0}}}

\def\vtheta{{\bm{\theta}}}

\def\vc{{\bm{c}}}

\def\vn{{\bm{n}}}

\def\vp{{\bm{p}}}
\def\vq{{\bm{q}}}

\def\vv{{\bm{v}}}

\def\vx{{\bm{x}}}
\def\vy{{\bm{y}}}
\def\vz{{\bm{z}}}

\def\mL{{\bm{L}}}
\def\mM{{\bm{M}}}

\DeclareMathAlphabet{\mathsfit}{\encodingdefault}{\sfdefault}{m}{sl}
\SetMathAlphabet{\mathsfit}{bold}{\encodingdefault}{\sfdefault}{bx}{n}

\def\sN{{\mathbb{N}}}

\newcommand{\R}{\mathbb{R}}

\DeclareMathOperator*{\argmin}{arg\,min}

\usepackage{hyperref}
\usepackage{url}
\usepackage{graphicx}
\usepackage[caption=false]{subfig}
\usepackage{capt-of}
\usepackage{xcolor}
\usepackage{wrapfig}
\usepackage{needspace}
\usepackage{listings}
\usepackage{booktabs}
\usepackage{tabularx}
\usepackage{pifont}
\usepackage{marvosym}
\usepackage{fontawesome5}
\usepackage{amsmath}
\usepackage{algorithm}
\usepackage{algpseudocode}

\hypersetup{hidelinks}

\newcommand{\xmark}{\text{\ding{55}}}
\newcommand{\deltagain}[1]{\textcolor[HTML]{9564DD}{\ensuremath{\mathbf{#1}}}}
\newcommand{\authoremail}[1]{\href{mailto:#1}{\texttt{#1}}}

\title{MaskFlow: Precise, Consistent and Seamless Regional Image Editing}

\author{
  Rui Xu$^{1,\dagger}$, Yang Yong$^{1}$, Shunzi Yang$^{2}$, Ruihao Gong$^{1,2,}$\textsuperscript{\Letter},
  Chengtao Lv$^{1,3}$ \\
  \multicolumn{1}{c}{
    $^{1}$ SenseTime Research \quad
    $^{2}$ Beihang University \quad
    $^{3}$ Nanyang Technological University
  } \\
  \multicolumn{1}{c}{
    $^{\dagger}$ \authoremail{chiaroair@gmail.com} \quad
    \textsuperscript{\Letter} \authoremail{gongruihao@buaa.edu.cn}
  } \\ \\
  \multicolumn{1}{c}{\faGlobe\ Project: \href{https://reychiaro.github.io/MaskFlow}{\texttt{https://reychiaro.github.io/MaskFlow}}}
}

\iclrfinalcopy

\begin{document}
\raggedbottom
\maketitle

\begin{figure}[htbp]
  \centering
  \vspace{-10pt}
  \includegraphics[width=1.0\linewidth]{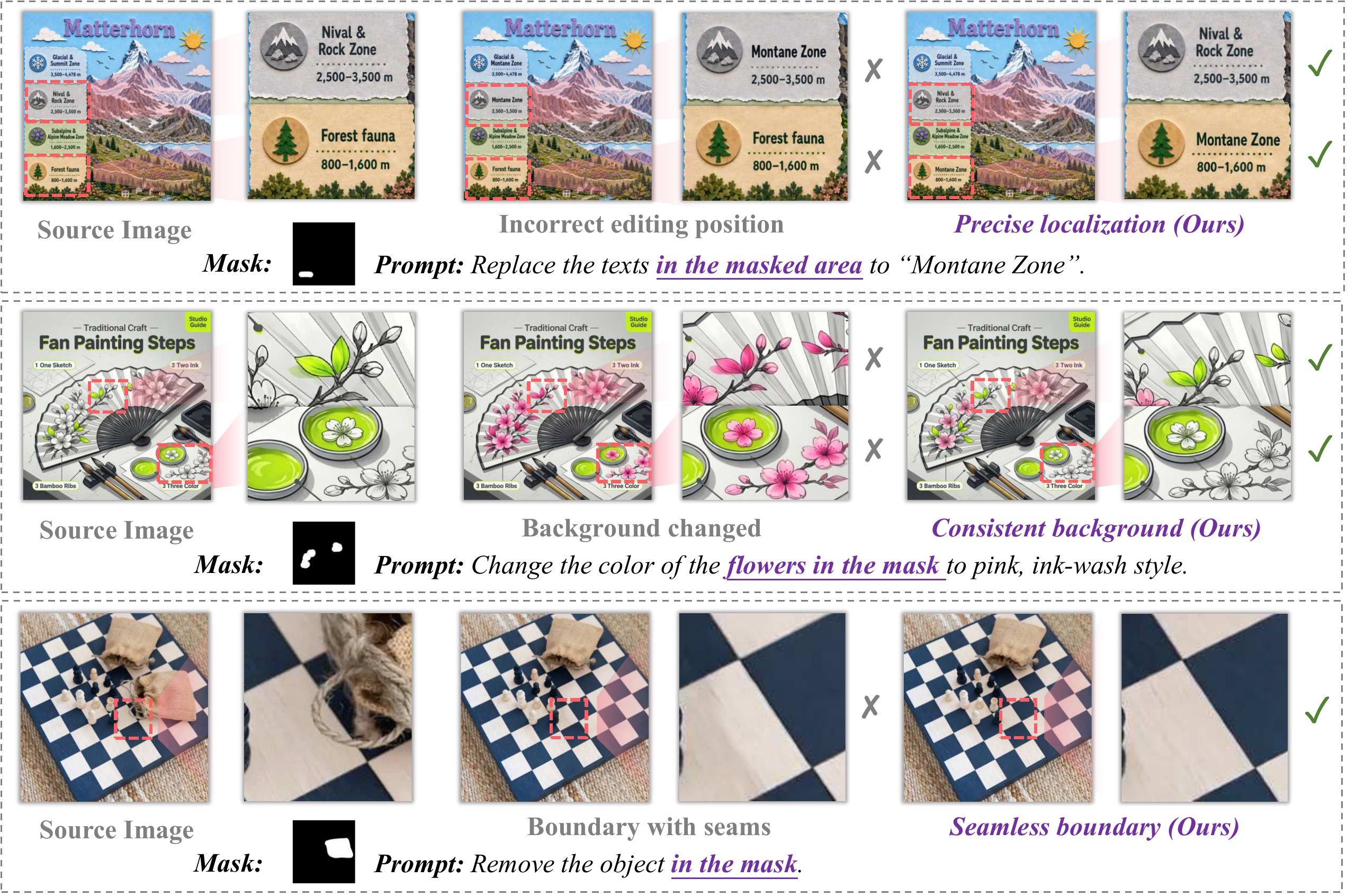}
  \vspace{-10pt}
  \caption{Qualitative comparison of the proposed \textbf{MaskFlow} with representative reference-based and inpainting methods. MaskFlow precisely localizes edits to masked regions, consistently preserves background content, and produces seamless transitions across mask boundaries.}
  \label{fig:teaser}
\end{figure}

\begin{abstract}
  Regional image editing has attracted considerable attention for its spatial controllability.
  Although instruction-based and mask-based editing methods can achieve strong semantic alignment, reliable regional control remains challenging, where an edit must be accurately localized and naturally integrated with the preserved context. 
  We propose \textbf{MaskFlow}, a training framework for precise localization, consistent background preservation, and seamless boundary transitions. MaskFlow incorporates the mask into the probability path and flow-matching objective, coordinating generation within the editable region with source preservation outside it. The proposed Soft-Poisson De-seaming module further refines the predicted vector field during both training and sampling to improve the smooth integration of the edited foreground with the preserved background. We also introduce MaskEdit-Benchmark for general scene and infographics editing, where prompts describe the desired edits without localization cues, leaving masks to specify the target regions.
  Experiments on natural scenes and infographic images demonstrate consistent improvements over competing methods in both quantitative and qualitative evaluations.
\end{abstract}

\section{Introduction}

Recent advances in diffusion and flow-matching models have substantially improved the fidelity and controllability of image generation~\citep{ddpm,ldm,flow-matching,dit,qwenimage}.
These developments have stimulated growing interest in conditioning mechanisms and controllable generation, particularly in regional image editing, which aims to modify regions specified by the user while preserving the remaining content~\citep{prompt2prompt,instructpix2pix,icedit}.

Existing approaches to regional image editing can be broadly categorized according to how the editable region is specified. Instruction-based methods identify the target content from textual instructions, whereas mask-based methods use explicit masks as spatial references.
Benefiting from advances in large language models and vision language models~\citep{clip,gpt4,gemini,bai2025qwen3}, instruction-based image editors have demonstrated strong semantic understanding and editing capabilities~\citep{gill,fireedit,smartedit,insightedit}, especially when the target can be described unambiguously.
However, as illustrated in~\Figref{fig:teaser}(a,b), specifying an exact target through language becomes cumbersome in complex scenes such as infographic images~\citep{bizgen,igenbench}. These images often contain repeated or visually similar elements, dense layouts, or targets whose locations are difficult to describe precisely. Such spatial ambiguity can substantially reduce localization accuracy.
Moreover, instruction-based methods commonly predict noise or vector fields over the entire spatial representation without explicit regional constraints~\citep{qwenimage,fluxkontext,bagel}. Consequently, they may unintentionally alter background content that should remain unchanged.

Mask-based methods alleviate spatial ambiguity by allowing users to specify the editable region explicitly~\citep{blended-latent-diffusion,imageneditor,smartbrush,powerpaint,refineanything}.
Given a source image and a mask that identifies the editable region, existing methods typically encode the mask as an additional image condition and process it in a manner similar to the source image. Common strategies introduce mask information through shared attention layers~\citep{attention}, additional conditioning branches~\citep{controlnet,t2i-adapter}, or parameter efficient fine-tuning~\citep{lora,icedit}.
Despite their improved localization capability, the mask-based methods continue to face challenges in edit fidelity and image integration. Generating the masked region independently can lead to semantic or stylistic drift~\citep{ldm,smartbrush}, which reduces the reliability of the edited content, may creating visible seams along the mask boundary and produce suboptimal results, as shown in~\Figref{fig:teaser}(c). Taken together, \textit{these limitations call for a regional editing framework that provides precise localization, consistent background preservation, and seamless boundary transitions}.

To address these challenges, we propose \textbf{MaskFlow}, a training and inference framework for regional image editing. Given a source image, an editing instruction, and a mask of arbitrary shape, MaskFlow firstly samples noise and predict the vector field conditioned on source and mask images. Then, it calculates the clean image and conducts Soft-Poisson De-seaming to produce a refined vector field. The refined vector is used to update the next noisy latent features.
Specifically, the proposed MaskFlow solves the problems through the following designs.
\textbf{(1)}
The localization capability is enhanced by encouraging the model to identify the editable regions from the masks via heuristic training on the constructed dataset.
\textbf{(2)}
The background is preserved with the proposed Siamese Probability Paths that recover the source image background while guide foreground editing.
\textbf{(3)}
The seamless boundary is refined by the designed Soft-Poisson De-seaming module that eliminates the visible discontinuities between the foreground and background by solving Poisson equation iteratively.
As demonstrated in~\Figref{fig:teaser}(a), MaskFlow accurately replaces all text instances specified by the masks. \Figref{fig:teaser}(b) further shows that semantically related background content remains unchanged. The results in~\Figref{fig:teaser}(c) illustrates more natural transitions than methods that directly replace masked regions.
\textbf{(4)}
We introduce MaskEdit-Benchmark, a training and evaluation dataset for mask-based image editing, that pairs region masks with concise referential instructions, decoupling spatial localization from editing intent to better reflect users' preference for concise prompts once the editing region is specified. The dataset further covers infographic editing, using masks to specify local regions that are difficult to describe textually and supporting the evaluation of models' ability to jointly interpret spatial guidance and editing instructions.

\textbf{Contributions.} This work addresses localization, background preservation, and seamless transitions in regional image editing. Our contributions are summarized below.
\begin{itemize}
    \item We propose \textbf{MaskFlow}, a training and inference framework for regional image editors. Guided by the mask, the Siamese Probability Paths combines the foreground and background generation with clean contexts preview, and Soft-Poisson De-seaming refines the vector fields by iterative optimization at both training and sampling stages.
    \item We introduce MaskEdit-Benchmark, covering general scenes and infographics with masks and referential instructions that omit object identification and location descriptions. It supports training and evaluation of joint mask and text understanding for localized editing.
    \item Extensive qualitative and quantitative experiments demonstrate that the proposed approach performs the requested edits correctly while improving localization accuracy, background consistency, and boundary alignment.
\end{itemize}

\section{Related Work}

Diffusion models~\citep{ddpm,ldm} and flow-matching models~\citep{flow-matching,rectified-flow}, together with scalable transformer backbones~\citep{dit,sit}, provide the generative foundation for regional edit models. A model must accurately localize, keep the unmasked area unchanged, and integrate the generated foreground with its surrounding context. Existing work mainly addresses these requirements through mask-based editing and refinement at region boundaries.

\paragraph{Mask-based regional editing.}
Masks provide explicit spatial information that is difficult to express through language alone. Early diffusion inpainting methods either blend generated and source latents during denoising or condition the model on masked images, object shapes, and high resolution source content~\citep{blended-latent-diffusion,imageneditor,smartbrush}. PowerPaint~\citep{powerpaint} further uses task prompts to unify filling, insertion, removal, and shape guided inpainting. Recent generalist editors~\citep{mao2025ace++,li2026brushedit} encode masks within long context multimodal inputs or combine semantic target recognition with dedicated inpainting branches. Task specific methods~\citep{jiang2025smarteraser,wasserman2025paint} improve removal or insertion through guidance from masked regions and purpose built training pairs. RegionE~\citep{chen2025regione} and SpotEdit~\citep{qin2026spotedit} instead infer editable or stable tokens from intermediate features and selectively reuse background features.
Despite their different implementations, these approaches generally treat the mask or estimated region as a conditioning signal rather than a constraint on the underlying probability path.
In contrast, \textbf{MaskFlow} incorporates the user mask into both the probability path and the flow-matching objective. This design directly constrains edits with masks of arbitrary shapes while maintaining the source trajectory outside the mask.

\paragraph{Blending and refinement at boundaries.}
Classical Poisson editing~\citep{poisson} and its modified formulation~\citep{modified-poisson} reconstruct a pasted region by matching source gradients under destination boundary conditions. Deep Image Blending~\citep{zhang2020deep} augments this principle with learned content and style objectives. These methods can suppress abrupt pixel discontinuities, but they operate on a completed composite and may introduce color propagation or fail to repair semantic and structural incompatibility created during generation. Recent approaches move closer to generative models. \citet{follow-your-shape} uses trajectory divergence and feature injection to localize shape changes, whereas RefineAnything~\citep{refineanything} and PixPerfect~\citep{pixperfect} learn refiners that operate at boundaries or in pixel space. However, automatically inferred regions do not provide exact user control. Auxiliary losses and subsequent refinement also cannot directly correct intermediate generative decisions that produce an incompatible foreground. The proposed \textbf{MaskFlow} instead applies Soft-Poisson De-seaming at every sampling step to refine the predicted vector field, allowing boundary consistency to influence the entire generation trajectory.

\section{Preliminaries}
\label{sec:preliminaries}

\paragraph{Flow-matching.}
Define the initial noise distribution $p_1=\mathcal{N}(0,I)$ and target data distribution $p_1=p_{\mathrm{data}}$, flow-matching models~\citep{flow-matching} learn vector field to map the random variables $X_1 \sim p_1$ to $X_0 \sim p_0$. Rectified flow~\citep{rectified-flow} parameterizes the probability path using $x_{\sigma} = (1-\sigma) x_0 + \sigma \epsilon$ with continuous time-shifted $\sigma \in [0,1]$~\citep{qwenimage}. The conditional flow-matching loss can be written as
\begin{equation}
  \mathcal{L}_{\mathrm{CFM}}(\vtheta)=\mathbb{E}\left[ \| v_{\vtheta}(x_{\sigma},\sigma) - v_{\sigma}(x_0,\epsilon|x_{\sigma}) \|_2^2 \right],
  \label{eq:cfm}
\end{equation}
where $v_{\vtheta}(x_0,\epsilon|x_{\sigma})=\frac{d x_{\sigma}}{d \sigma}=\epsilon - x_0$ is the conditional velocity while $v_{\vtheta}(x_{\sigma},\sigma)$ is the predicted velocity.
Given pre-trained vector predictor $v_{\vtheta}(x_{\sigma},\sigma)$ and the inference timestep sequence $\{ T, T-1, \cdots, 1 \}$, the predicted clean target at timestep $t$ is
$\hat{x}_0^t = x_{\sigma_t} - \sigma_t v_{\vtheta} (x_{\sigma_t}, \sigma_t)$,
where the continuous $\sigma$ distribution is discretized by timesteps for iterative denoising.



\section{Methodology}

We present the proposed MaskFlow framework in~\Secref{sec:maskflow} and Soft-Poisson De-seaming (SPD) in~\Secref{sec:soft-poisson-deseaming}. A novel benchmark for mask-based image editing constructed by the data synthesis pipeline will be introduced in~\Secref{sec:data-pipeline}.

\subsection{MaskFlow}
\label{sec:maskflow}

\begin{figure}[t]
  \centering
  \includegraphics[width=0.85\textwidth]{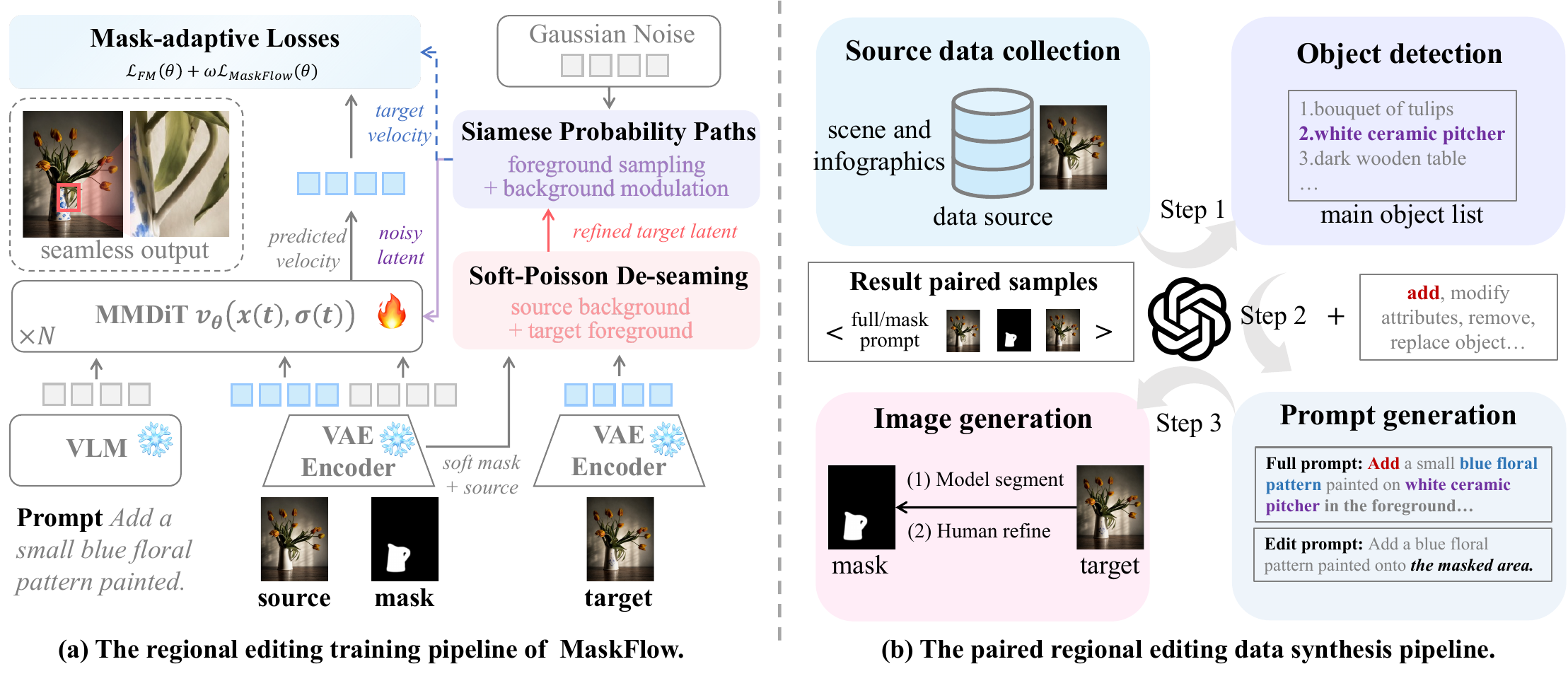}
  \caption{Overview of the proposed \textbf{MaskFlow} framework and data synthesis pipeline. \textbf{(a)} During training, the source image and mask are provided to the MMDiT for vector field prediction. \textbf{(b)} The three-stage data synthesis pipeline uses an VLM and an image generation model to produce \texttt{<prompt,source,mask,target>} samples.}
  \label{fig:arch-data_pipe}
\end{figure}

We now introduce the MaskFlow framework for training and inference. The proposed framework integrates freestyle mask in probability path as well as the supervision losses. Siamese Probability Paths (SPP) provide cleaner background context to guide foreground generation, while the MaskFlow loss strengthens supervision within the editable foreground region.

\paragraph{Siamese Probability Paths.} To begin with, we use same flow-matching notations in~\Secref{sec:preliminaries} with $\sigma$-based probability path $\vx_{\sigma} = (1-\sigma) \vx_0 + \sigma \epsilon$, where $\vx_0$ denotes the edited image and $\vx_1=\epsilon$ is the noise. To eliminate the unwilling background changes, we split the image into the editable region and the background context region. Then, the siamese probability paths are generated with
\begin{equation}
  \vx_{\sigma} = \vx_{\sigma}^{\widetilde{\mM}} + \vx_{\sigma,\gamma}^{1-\widetilde{\mM}},
  \label{eq:spp}
\end{equation}
where $\vx_{\sigma}^{\widetilde{\mM}} = \widetilde{m} \odot ((1-\sigma) \vx_0^{\star} + \sigma \epsilon)$ is the mask-cropped rectified flow probability path and $\vx_0^{\star}$ is the boundary refined image using SPD. The second item $\vx_{\sigma,\gamma}^{1-\widetilde{\mM}}$ is parameterized with $\gamma > 1$ using the background context coming from the source image:
\begin{equation}
  \vx_{\sigma,\gamma}^{1-\widetilde{\mM}}=(1-\widetilde{m}) \odot ((1-\sigma^{\gamma}) \vx_S + \sigma^{\gamma} \epsilon).
  \label{eq:spp-gamma}
\end{equation}
As $\gamma > 1$ and $\sigma \in [0,1]$, the background has more cleaner structure and semantic than the editable region for better guiding the generation process.

\paragraph{MaskFlow Loss.} With the siamese probability paths, the target vector field is
$\vv_{\sigma}=\frac{\mathrm{d} \vx_{\sigma}}{\mathrm{d} \sigma} = \widetilde{m} \odot (\epsilon - \vx_0^{\star}) + (1-\widetilde{m}) \odot \gamma \sigma^{\gamma-1} (\epsilon - \vx_S)$. Then the conditional flow-matching loss~\Eqref{eq:cfm} can be conducted on this target and model predicted vector.
However, the proposed MaskFlow is designed to suit masks with different shapes and sizes provided by users. Supervised solely on the vector field defined on the whole image has weak signals on small size masks. To overcome the problem, we design a mask-adaptive weighted loss to add additional supervision on the masked area
\begin{equation}
  \mathcal{L}_{\mathrm{MaskFlow}}(\vtheta) = \mathbb{E}\left[ \frac{HW}{\sum_{H,W} \widetilde{\mM}}\widetilde{m}\odot\| (\vv_{\vtheta}(\vx_{\sigma},\sigma) - \vv_{\sigma}) \|_2^2 \right],
  \label{eq:loss-maskflow}
\end{equation}
the loss weight is sample-wise and adaptive to the area of soft mask, balancing the contribution between small and large masks. The overall supervision loss is $\mathcal{L}(\vtheta) = \mathcal{L}_{\mathrm{CFM}}(\vtheta) + \omega_{\widetilde{\mM}} \mathcal{L}_{\mathrm{MaskFlow}}(\vtheta)$,
where $\omega_{\widetilde{\mM}}$ is a loss weight to adjust the scale of MaskFlow loss.

\paragraph{Training and Inference.} As shown in~\Figref{fig:arch-data_pipe}(a), training begins by applying SPD to the edited target to obtain the boundary-refined image $\vx_0^{\star}$. The noisy input is then constructed using the siamese probability paths in~\Eqref{eq:spp}, with the source background following~\Eqref{eq:spp-gamma}. Conditioned on the source image, mask, and editing instruction, the model predicts the vector field and is optimized with the overall supervision loss $\mathcal{L}(\vtheta)$.
During inference, at timestep $t$, the model predicts $\vv_{\vtheta}(\vx_{\sigma_t},\sigma_t,\vc)$ under the given conditions $\vc$. The predicted clean sample is computed as $\widehat{\vx}_0^t=\vx_{\sigma_t} - \sigma_t \vv_{\vtheta} + C(\sigma_t,\gamma)$, where the bias correction $C(\sigma_t,\gamma)$ is derived in~\Secref{apx:clean-sample-prediction}. SPD then refines this estimate to obtain $\widehat{\vx}_{0}^{t,\star}$, and the correction $\frac{1}{\sigma_t}(\widehat{\vx}_0^{t}-\widehat{\vx}_{0}^{t,\star})$ is added to the predicted vector field to produce $\vv_{\vtheta}^{\star}$ for the next denoising step. The complete algorithms are provided in~\Secref{apx:algrithms}. Both stages thus rely on Soft-Poisson De-seaming to integrate the edited foreground with the source background, and we will introduce it in the following section.

\subsection{Soft-Poisson De-seaming}
\label{sec:soft-poisson-deseaming}

\begin{figure}[t]
  \centering
  \includegraphics[width=0.90\textwidth]{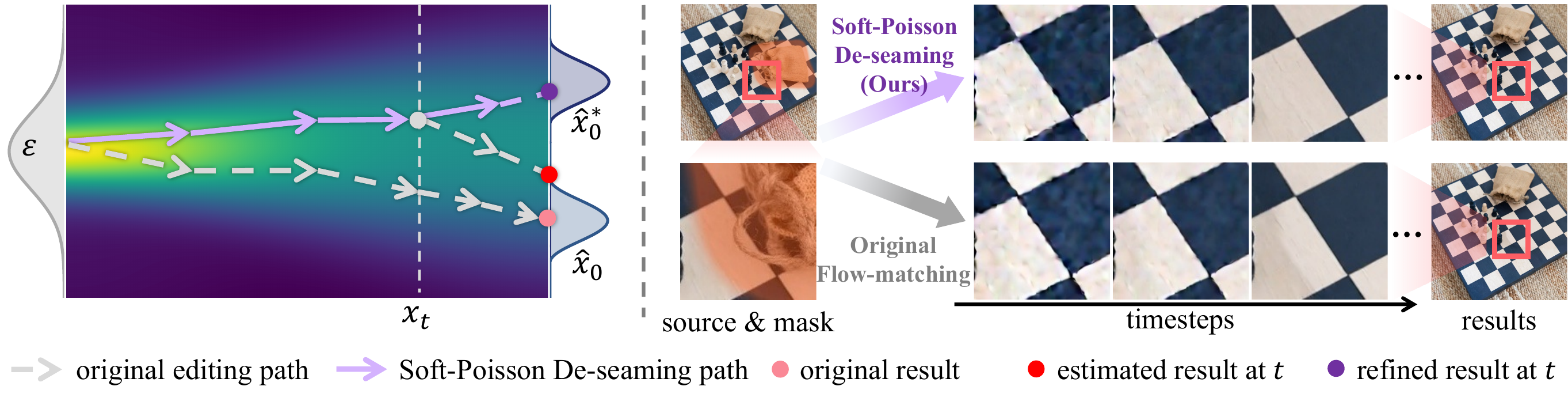}

  \caption{Probability paths for Soft-Poisson De-seaming (SPD) and conventional flow-matching. \textbf{Left.} SPD corrects the vector field at time $t$ and moves the sample toward the seamless distribution. \textbf{Right.} The upper result from our method contains fewer boundary artifacts.}
  \label{fig:method-poisson}
\end{figure}

This section introduces \textbf{Soft-Poisson De-seaming} (SPD), which refines the predicted vector field at every training and sampling step through a unified gradient-domain objective as shown in~\Figref{fig:method-poisson}.
Starting from the binary mask $\mM\in\{0,1\}^{H\times W}$ where $\mM_{i,j}=1$ denotes the editable token, the soft mask is obtained by Gaussian blur $\widetilde{\mM}=\operatorname{GaussianBlur}(\mM)$. The expanded editable region is $\widetilde{\Omega} =\left\{ \vp=(i,j)\in\{1,\ldots,H\}\times\{1,\ldots,W\} \;\middle|\; \widetilde{\mM}_{i,j}>0 \right\}$. Given the source image $\vx_S$ and a candidate image $\vy$ (such as the edited image and the predicted clean image), the target of SPD is to paste the masked region from $\vy$ to $\vx_S$ with a soft and seamless manner especially in the boundary.

Note that the tokens in the boundary between the editable and background areas is covered by the soft mask, whose values are modulated by the soft mask $\widetilde{\mM}$. Let $\widetilde m(\vp)=\widetilde{\mM}_{i,j}$ for $\vp=(i,j)$.
We obtain the boundary refined image by solving
\begin{equation}
  \begin{aligned}
  &\vz^{\star}=\argmin_{\vz}\;
  \int_{\widetilde{\Omega}}
    \widetilde m(\vp)
    \left\lVert
      \nabla_{\vp}\vz(\vp)-\nabla_{\vp}\vy(\vp)
    \right\rVert_F^2
  \,\mathrm{d}\vp
  +\int_{\widetilde{\Omega}}
    \widetilde m(\vp)
    \left\lVert\vz(\vp)-\vy(\vp)\right\rVert_2^2
  \,\mathrm{d}\vp \\
  &\qquad+\int_{\widetilde{\Omega}}
    \left(1-\widetilde m(\vp)\right)
    \left\lVert\vz(\vp)-\vx_S(\vp)\right\rVert_2^2
  \,\mathrm{d}\vp
  \qquad\mathrm{s.t.}\quad
  \vz\vert_{\partial\widetilde{\Omega}}
  =\vx_S\vert_{\partial\widetilde{\Omega}},
  \end{aligned}
  \label{eq:spd}
\end{equation}
where the final solved $\vz^{\star}$ is the blended image with refined seamless boundary.
The first term weights gradient alignment by the soft mask to preserve structure while reducing its strength in the transition region. The second term anchors the refined field to the regions with large soft-mask values. The third term progressively restores the source feature as the soft-mask value decreases. Together with the boundary condition, these terms form a continuous transition from the generated foreground to the preserved background within a single optimization objective.


In practice, we solve \Eqref{eq:spd} with Euler--Lagrange equation and discretize it for efficient numerical computation. Let $\operatorname{N}(\vp)$ be the set of neighboring pixels of $\vp\in\widetilde{\Omega}$, and let $\widetilde m_{\vp}$ denote the corresponding soft-mask value. Define
$w_{\vp\vq}=(\widetilde m_{\vp}+\widetilde m_{\vq})/2$ and $J_{\vp}=\sum_{\vq\in\operatorname{N}(\vp)}w_{\vp\vq}$,
the discrete equation using Jacobi iteration is
\begin{equation}
  \resizebox{0.98\linewidth}{!}{$\displaystyle
  \vz_{\vp}^{(k+1)}=\frac{1}{J_{\vp}+1}\Big[
  \sum_{\scriptscriptstyle{\vq\in\operatorname{N}(\vp)\cap\widetilde{\Omega}}}w_{\vp\vq}\vz_{\vq}^{(k)}
  +\sum_{\scriptscriptstyle{\vq\in\operatorname{N}(\vp)\backslash\widetilde{\Omega}}}w_{\vp\vq}\vx_{S,\vq}
  +\sum_{\scriptscriptstyle{\vq\in\operatorname{N}(\vp)}}w_{\vp\vq}\left(\vy_{\vp}-\vy_{\vq}\right)
  +\widetilde m_{\vp}\vy_{\vp}
  +\left(1-\widetilde m_{\vp}\right)\vx_{S,\vp}
  \Big].
  $}
  \label{eq:spd-jacobi}
\end{equation}
The complete derivation is provided in supplementary \Secref{apx:poisson-soft}. In practice, $\vz$ is updated using relaxed Jacobi iteration. After $K$ iterations, the refined image is approximated by $\vy_{\vp}^{\star}\approx\vz_{\vp}^{(K)}$.

\subsection{MaskEdit Benchmark}
\label{sec:data-pipeline}

We introduce MaskEdit-Benchmark in this section to support training and evaluation by decoupling spatial localization from edit instructions. It includes editing samples from both general scenes and infographics.
As shown in \Figref{fig:arch-data_pipe}(b), the pipeline contains object detection, prompt generation, and image generation.
Given source images containing multiple concepts and potentially editable objects, \textbf{object detection} identifies the salient objects in each image. As there may be many visually insignificant objects whose editing provides limited value for training, we prompt a vision language model to identify the primary objects in each source image.
The \textbf{prompt generation} stage produces two instructions for regional editing. The first is a complete instruction that specifies the editing operation, target position, and desired result. The second replaces the explicit position description with a demonstrative expression, encouraging the model to obtain localization information from the mask during training.
The final \textbf{image generation} stage synthesizes target images and constructs their masks. We use the complete instructions with detailed positions to generate the target images. SAM~\citep{sam} then provides an initial segmentation of each target region and then the human annotators refine these regions and create masks of arbitrary shapes that resemble practical user inputs.
Dataset composition and counting conventions are detailed in~\Secref{apx:data-analysis}.

\section{Experiments}

\begin{table}[htb]
\centering
\captionsetup{position=top}
\captionsetup[subtable]{position=top}
\caption{Editing performance and background consistency on mask-based editing tasks. \textbf{Bold} is the best and \underline{underline} is the second best within each task. $\Delta$ denotes the gains from our framework.}
\label{tab:baseline-comparisons}
\subfloat[General scene image editing\label{tab:baseline-scene}]{%
\resizebox{0.95\textwidth}{!}{%
  \begin{tabular}{l|ccccccc|cc}
    \toprule
    \multicolumn{1}{c}{} & \multicolumn{7}{|c}{Edit Performance} & \multicolumn{2}{|c}{Background Consistency} \\
    \midrule
    Model & CLIP-I $\uparrow$ & CLIP-C $\uparrow$ & DINO $\uparrow$ & FID $\downarrow$ & PSNR $\uparrow$ & SSIM $\uparrow$ & LPIPS $\downarrow$ & $\mathrm{SSIM}_{\mathrm{bg}} \uparrow$ & $\mathrm{LPIPS}_{\mathrm{bg}} \downarrow$ \\
    \midrule
    Gemini 3~\citep{gemini}  & 0.9715 & 0.8726 & \underline{0.9533} & \underline{20.32} & \underline{21.55} & 0.7184 & 0.1990 & 0.8606 & 0.1048 \\
    GPT Image 2~\citep{gpt4} & \underline{0.9728} & 0.9111 & 0.9361 & 24.16 & 16.74 & 0.5039 & 0.2594 & 0.6074 & 0.1819 \\
    \midrule
    RefineAnything~\citep{refineanything} & 0.9625 & \underline{0.9321} & 0.9060 & 26.35 & 20.85 & \textbf{0.7435} & 0.1856 & \textbf{0.9999} & \textbf{0.0001} \\
    RegionE~\citep{chen2025regione} & 0.96545 & 0.8637 & 0.9314 & 25.89 & 20.55 & 0.7065 & 0.2465 & 0.8690 & 0.1378 \\
    SpotEdit~\citep{qin2026spotedit} & 0.9199 & 0.8807 & 0.8899 & 56.80 & 10.83 & 0.3049 & 0.6113 & 0.4161 & 0.5122 \\
    QwenImage-Inpaint~\citep{lugmayr2022repaint} & 0.9714 & 0.9226 & 0.9403 & 23.07 & 21.41 & 0.7380 & 0.1489 & \underline{0.9161} & \underline{0.0440} \\
    \midrule
    BAGEL-7B-MoT~\citep{bagel} & 0.9436 & 0.8969 & 0.9056 & 27.69 & 20.30 & 0.7201 & 0.2488 & 0.8810 & 0.1382 \\
    HiDream-O1-Image~\citep{hidream} & 0.9116 & 0.8329 & 0.8423 & 45.16 & 13.09 & 0.3732 & 0.5623 & 0.4887 & 0.4544 \\
    \midrule
    FLUX.2-dev~\citep{fluxkontext} & 0.9546 & 0.8932 & 0.9138 & 29.96 & 19.20 & 0.6473 & 0.2296 & 0.7911 & 0.1246 \\
    + MaskFlow (ours) & 0.9669 & 0.9136 & 0.9291 & 24.11 & 20.94 & 0.7397 & \underline{0.1216} & \textbf{0.9999} & \textbf{0.0001} \\
    \deltagain{\Delta (\%)} & \deltagain{+1.29} & \deltagain{+2.28} & \deltagain{+1.67} & \deltagain{+19.53} & \deltagain{+9.06} & \deltagain{+14.27} & \deltagain{+47.04} & \deltagain{+26.41} & \deltagain{+100.0} \\
    \midrule
    QwenImage-2511~\citep{qwenimage} & 0.9455 & 0.8672 & 0.9103 & 29.83 & 18.80 & 0.6208 & 0.2804 & 0.7425 & 0.1793 \\
    + MaskFlow (ours) & \textbf{0.9790} & \textbf{0.9408} & \textbf{0.9579} & \textbf{18.44} & \textbf{21.95} & \underline{0.7424} & \textbf{0.1210} & \textbf{0.9999} & \textbf{0.0001} \\
    \deltagain{\Delta (\%)} & \deltagain{+3.54} & \deltagain{+8.49} & \deltagain{+5.23} & \deltagain{+38.18} & \deltagain{+16.76} & \deltagain{+19.59} & \deltagain{+56.85} & \deltagain{+34.68} & \deltagain{+100.0} \\
    \bottomrule
  \end{tabular}%
}
}

\subfloat[Infographics image editing\label{tab:baseline-infographic}]{%
\resizebox{0.95\textwidth}{!}{%
  \begin{tabular}{l|ccccccc|cc}
    \toprule
    \multicolumn{1}{c}{} & \multicolumn{7}{|c}{Edit Performance} & \multicolumn{2}{|c}{Background Consistency} \\
    \midrule
    Model & CLIP-I $\uparrow$ & CLIP-C $\uparrow$ & DINO $\uparrow$ & FID $\downarrow$ & PSNR $\uparrow$ & SSIM $\uparrow$ & LPIPS $\downarrow$ & $\mathrm{SSIM}_{\mathrm{bg}} \uparrow$ & $\mathrm{LPIPS}_{\mathrm{bg}} \downarrow$  \\
    \midrule
    Gemini 3~\citep{gemini} & 0.9806 & 0.8874 & 0.9581 & 26.12 & 19.31 & 0.8129 & 0.1165 & 0.9005 & 0.0553 \\
    GPT Image 2~\citep{gpt4} & 0.9558 & 0.8693 & 0.9508 & 37.81 & 14.85 & 0.6367 & 0.3390 & 0.6824 & 0.2942 \\
    \midrule
    RefineAnything~\citep{refineanything} & 0.9846 & 0.8836 & 0.9595 & 22.36 & 19.96 & 0.8509 & \underline{0.0818} & \underline{0.9998} & \underline{0.0002} \\
    RegionE~\citep{chen2025regione} & 0.9622 & 0.8449 & 0.9278 & 37.76 & 17.38 & 0.8044 & 0.1767 & 0.8972 & 0.1133 \\
    SpotEdit~\citep{qin2026spotedit} & 0.9729 & 0.8583 & 0.9527 & 27.10 & 18.79 & 0.8228 & 0.1342 & 0.9164 & 0.0686 \\
    QwenImage-Inpaint~\citep{lugmayr2022repaint} & \underline{0.9871} & \underline{0.8973} & \underline{0.9686} & \underline{20.80} & \underline{20.49} & \underline{0.8534} & 0.0875 & 0.9620 & 0.0193 \\
    \midrule
    BAGEL-7B-MoT~\citep{bagel} & 0.9662 & 0.8536 & 0.9464 & 31.85 & 17.99 & 0.8123 & 0.1582 & 0.9113 & 0.0919 \\
    HiDream-O1-Image~\citep{hidream} & 0.9303 & 0.8324 & 0.8722 & 61.73 & 12.16 & 0.5580 & 0.4368 & 0.6188 & 0.3844 \\
    \midrule
    FLUX.2-dev~\citep{fluxkontext} & 0.9689 & 0.8696 & 0.9388 & 36.06 & 16.40 & 0.7373 & 0.1588 & 0.8041 & 0.1037 \\
    + MaskFlow (ours) & 0.9720 & 0.8750 & 0.9578 & 25.37 & 19.51 & 0.8444 & 0.0939 & \textbf{0.9999} & \textbf{0.0001} \\
    \deltagain{\Delta (\%)} & \deltagain{+0.32} & \deltagain{+0.62} & \deltagain{+2.02} & \deltagain{+29.65} & \deltagain{+18.96} & \deltagain{+14.53} & \deltagain{+40.87} & \deltagain{+24.36} & \deltagain{+100.0} \\
    \midrule
    QwenImage-2511~\citep{qwenimage} & 0.9693 & 0.8765 & 0.9395 & 37.95 & 17.49 & 0.7921 & 0.1387 & 0.8710 & 0.0805 \\
    + MaskFlow (ours) & \textbf{0.9903} & \textbf{0.9137} & \textbf{0.9758} & \textbf{17.20} & \textbf{21.10} & \textbf{0.8575} & \textbf{0.0725} & \textbf{0.9999} & \textbf{0.0001} \\
    \deltagain{\Delta (\%)} & \deltagain{+2.17} & \deltagain{+4.24} & \deltagain{+3.86} & \deltagain{+54.68} & \deltagain{+20.64} & \deltagain{+8.26} & \deltagain{+47.73} & \deltagain{+14.81} & \deltagain{+100.0} \\
    \bottomrule
  \end{tabular}%
  }
  }
  
\end{table}

\subsection{Experimental settings}
\label{sec:experimental-settings}



\textbf{Implementation details.} We use QwenImage-2511~\citep{qwenimage} as the default backbone and additionally evaluate MaskFlow on FLUX.2-dev~\citep{fluxkontext}. We train attention LoRA modules~\citep{lora,attention} for 5K steps with rank 256 using the Prodigy optimizer~\citep{mishchenko2023prodigy} and an adaptive learning rate of $1.0$. We set the MaskFlow loss weight to $\omega_{\widetilde{\mM}}=2.0$ and the background path exponent to $\gamma=1.5$. Soft-Poisson De-seaming uses $K=50$ Jacobi iterations~\Eqref{eq:spd-jacobi}. During inference, we use $50$ sampling steps and a classifier-free guidance scale of $4.0$~\citep{ho2022classifier}.

\textbf{Evaluation metrics.} We evaluate editing quality and background preservation on the general scene and infographics subsets of MaskEdit-Benchmark. CLIP-I~\citep{clip} and DINO~\citep{oquab2023dinov2} measure semantic and structural similarity between edited results and target images. CLIP-C~\citep{clip} evaluates the foreground semantic similarity. FID~\citep{heusel2017gans} evaluates distributional fidelity, while PSNR~\citep{psnr}, SSIM~\citep{ssim}, and LPIPS~\citep{zhang2018unreasonable} assess pixel, structural, and perceptual fidelity.
Background consistency is measured against the source image in unmasked regions using $\mathrm{SSIM}_{\mathrm{bg}}$ and $\mathrm{LPIPS}_{\mathrm{bg}}$.

\begin{figure}[htb]
  \centering
  \subfloat[Qualitative comparisons on general scene image editing task\label{fig:baseline-scene}]{%
    \includegraphics[width=0.90\linewidth]{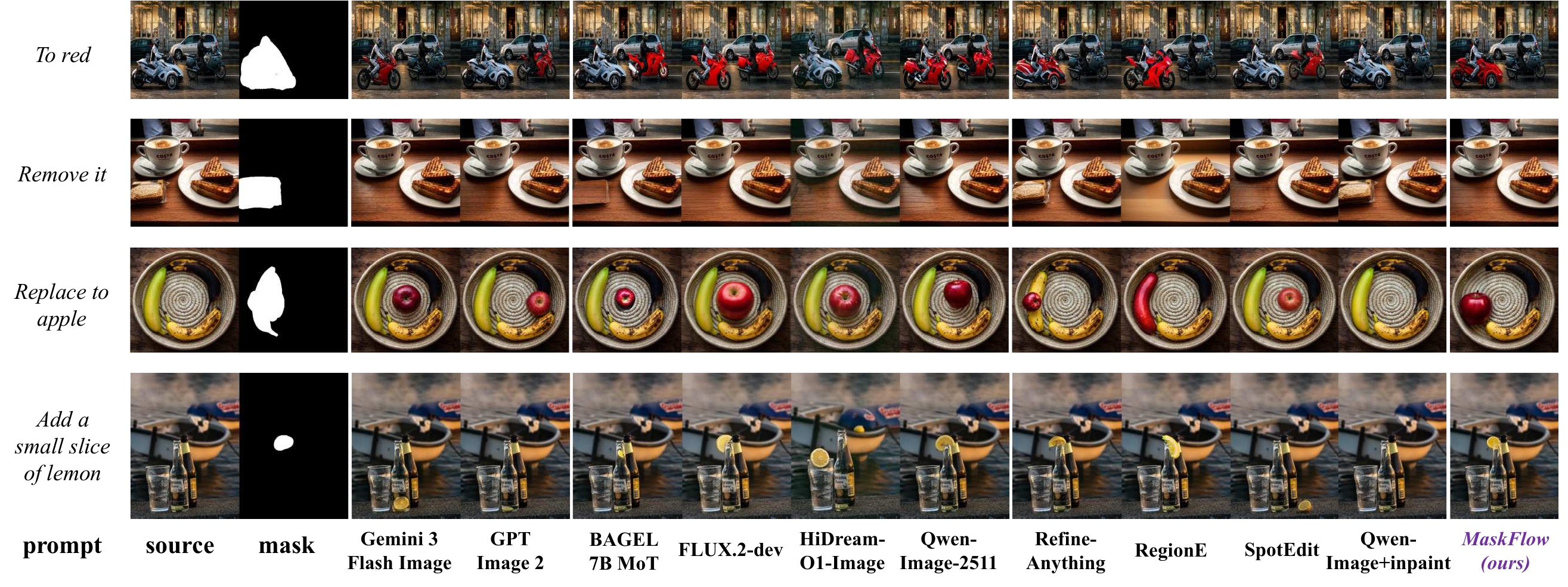}%
  }

  \subfloat[Qualitative comparisons on infographics image editing task\label{fig:baseline-infographic}]{%
    \includegraphics[width=0.90\linewidth]{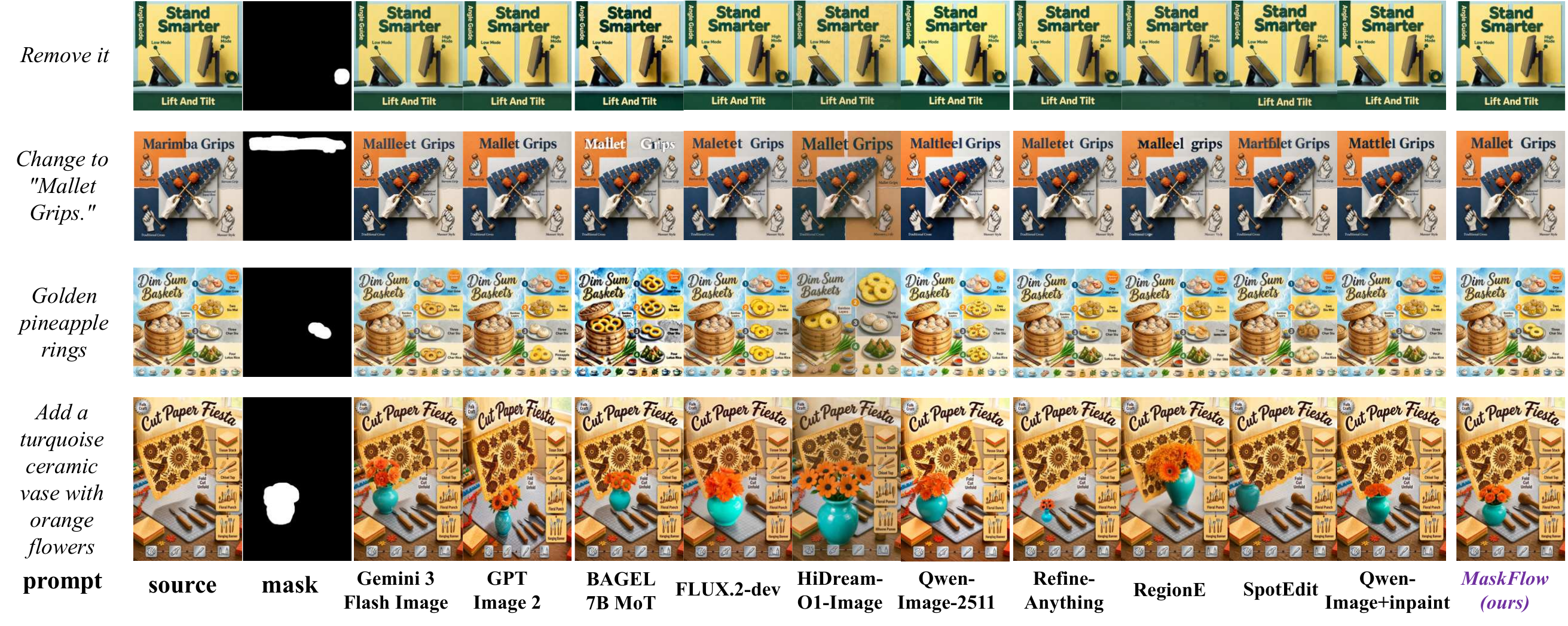}%
  }
  \caption{Qualitative comparisons on general scenes and infographics. MaskFlow accurately edits the masked regions while preserving surrounding content and maintaining natural visual transitions. Competing methods exhibit inaccurate localization, unintended background changes, or inconsistent edited content. Please zoom in for details.}
  \label{fig:baseline-comparisons}
\end{figure}

\subsection{Image editing comparisons}

\textbf{Quantitative analysis.}
Table~\ref{tab:baseline-comparisons} shows that MaskFlow delivers strong editing quality together with consistent background preservation across both tasks. On general scenes, MaskFlow with QwenImage-2511 leads in semantic similarity, structural similarity, distributional fidelity, PSNR, and perceptual fidelity, while remaining competitive with the strongest SSIM result. On infographics, it achieves the best results across all reported metrics, demonstrating its effectiveness in images with dense layouts and intertwined visual and textual content. Both MaskFlow variants attain the optimal reported background scores, showing that high-quality edits can be achieved while faithfully retaining unmasked content.
The improvements also hold across backbones. Applying MaskFlow to either FLUX.2-dev or QwenImage-2511 improves every reported metric on both tasks, with particularly substantial gains in distributional and perceptual fidelity. These consistent improvements demonstrate that the framework strengthens regional editing across different base models and image domains.

\textbf{Qualitative analysis.}
\Figref{fig:baseline-scene} illustrates MaskFlow's precise regional control across modification, removal, replacement, and addition. In scenes with similar objects, competing methods may modify the wrong instance or spread the edit beyond the mask. MaskFlow recolors the selected motorcycle and replaces the selected banana with an apple while preserving neighboring objects. It also removes the masked item from the tabletop and places a lemon slice at the specified location with natural integration into the scene.
\Figref{fig:baseline-infographic} highlights the same advantages in more demanding layouts. MaskFlow accurately updates the masked heading, replaces the selected food item, and inserts the requested vase while retaining surrounding text and illustrations. Other methods exhibit misspelled text, changes to unrelated elements, or inaccurate placement. Together, these examples show that MaskFlow combines reliable instruction following with precise localization, background preservation, and coherent visual integration. 
More visual comparisons are provided in Appendix~\ref{apx:visualization}.

\subsection{Ablation studies}

\begin{figure}[t]
\centering
\begin{minipage}[c]{0.46\textwidth}
  \centering
  \setlength{\belowcaptionskip}{4pt}
  \captionof{table}{Ablation of the proposed modules.}
  \label{tab:abl-method}
  \resizebox{\linewidth}{!}{%
    \begin{tabular}{ccc|ccc}
      \toprule
      SPD & SPP & MFL & CLIP-I $\uparrow$ & CLIP-C $\uparrow$ & FID $\downarrow$ \\
      \midrule
      - & - & - & 0.9726 & 0.9206 & 21.46  \\
      $\checkmark$ & - & - & 0.9775 & 0.9377 & 19.06 \\
      $\checkmark$ & $\checkmark$ & - & 0.9788 & 0.9380 & 18.91 \\
      $\checkmark$ & $\checkmark$ & $\checkmark$ & \textbf{0.9790} & \textbf{0.9408} & \textbf{18.44} \\
      \bottomrule
    \end{tabular}%
  }
\end{minipage}\hfill
\begin{minipage}[c]{0.51\textwidth}
  \centering
  \setlength{\belowcaptionskip}{4pt}
  \includegraphics[width=\linewidth]{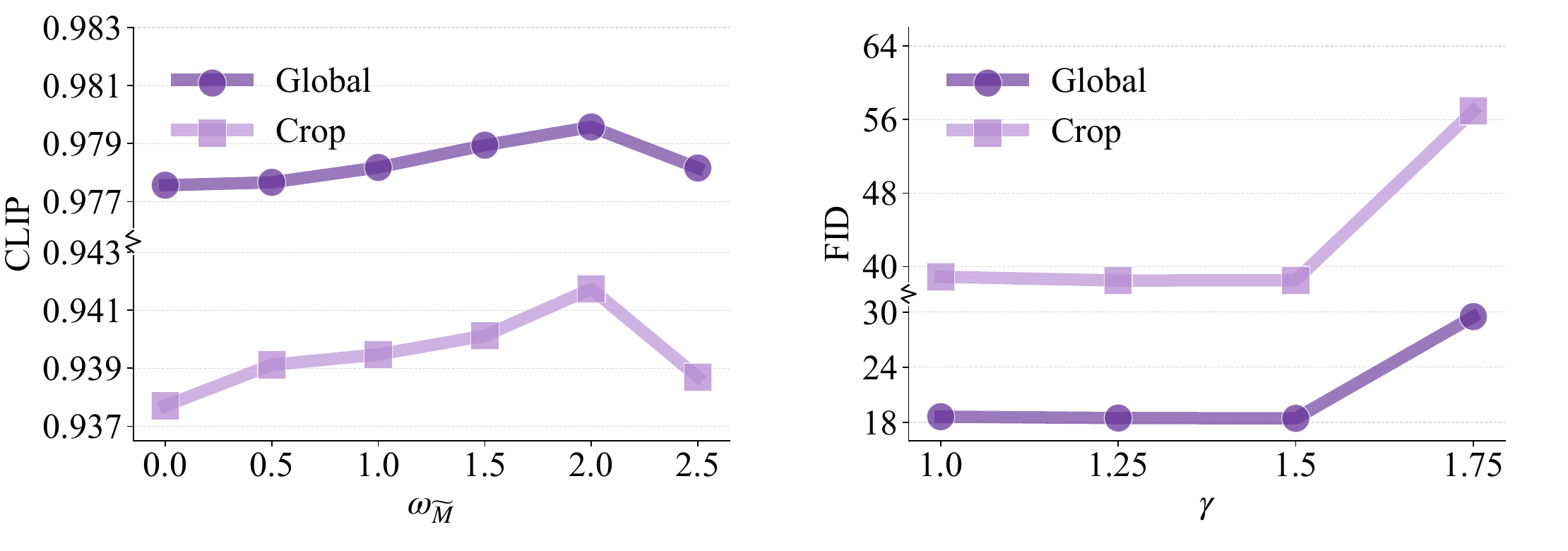}
  \captionof{figure}{Parameter sensitivity of MaskFlow.}
  \label{fig:abl-param}
\end{minipage}
\par\medskip
  \centering
  \captionof{table}{Effect of \textit{additional} textual position descriptions on regional image editing given the masks.}
  \label{tab:abl-precision}
  \setlength{\tabcolsep}{3pt}
  \begin{tabular}{c|ccccccc}
    \toprule
    Additional Position & CLIP-I $\uparrow$ & DINO $\uparrow$ & FID $\downarrow$ & PSNR $\uparrow$ & SSIM $\uparrow$ & LPIPS $\downarrow$ \\
    \midrule
    $\checkmark$ & 0.9779 & 0.9567 & 17.81 & 21.08 & 0.6993 & 0.1680 \\
    $\xmark$ \textbf{(Ours)} & \textbf{0.9798} & \textbf{0.9595} & \textbf{17.18} & \textbf{21.27} & \textbf{0.7001} & \textbf{0.1664} \\
    \bottomrule
  \end{tabular}
\end{figure}

\textbf{Effectiveness of the proposed modules.}
Table~\ref{tab:abl-method} evaluates the progressive addition of Soft-Poisson De-seaming (SPD), Siamese Probability Paths (SPP), and MaskFlow loss (MFL). The baseline is QwenImage-2511 fine-tuned on the same data for the same number of training steps as the other configurations. Introducing SPD yields the improvement over this baseline, substantially increasing foreground CLIP-C while improving CLIP-I and reducing FID. Adding SPP provides further gains across all three metrics. Incorporating MFL further improves foreground semantic similarity, as reflected in CLIP-C, and produces the best results across all three metrics. This progression supports the complementary roles of boundary refinement, coordinated foreground and background trajectories, and focused supervision of editable regions.

\begin{figure}[tbp]
  \centering
  \includegraphics[width=0.75\textwidth]{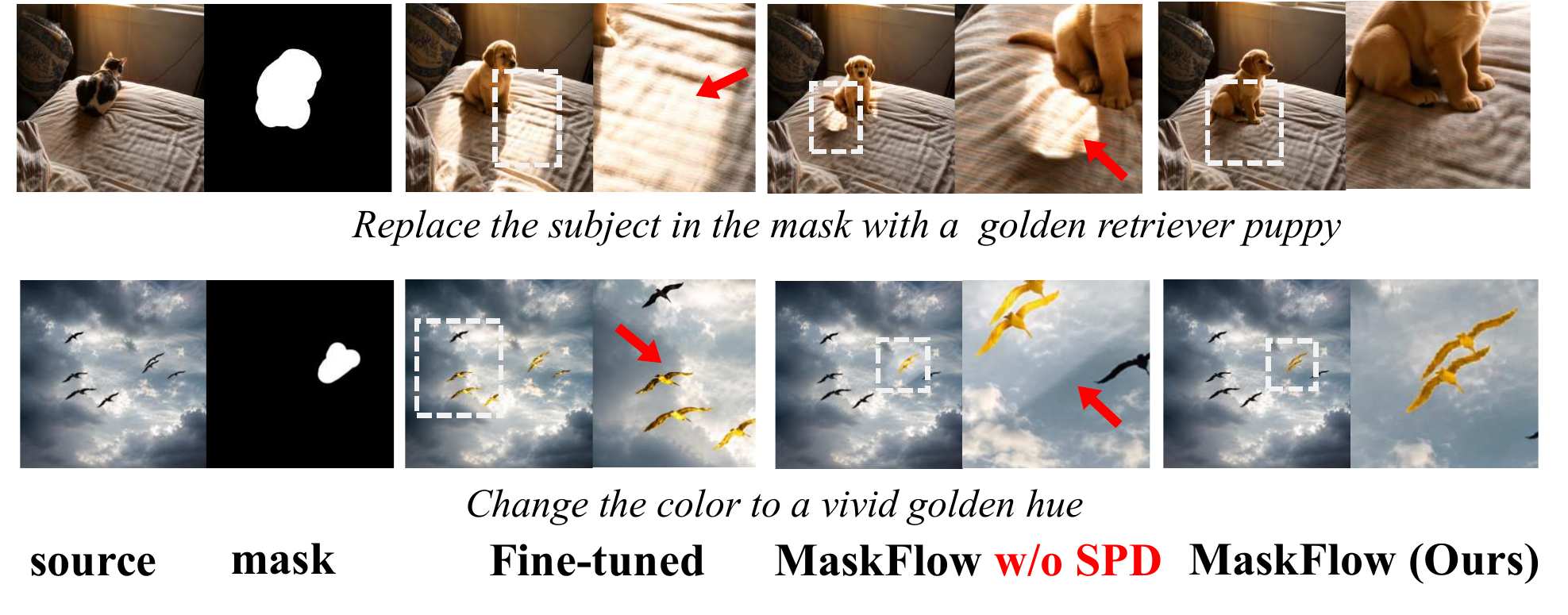}
  \caption{Qualitative ablation comparing the fine-tuned baseline, MaskFlow without SPD, and the full framework. Red arrows indicate visible errors. Please zoom in for details.}
  \label{fig:abl-method}
\end{figure}

The visual ablation in~\Figref{fig:abl-method} further clarifies these roles. The fine-tuned baseline alters the surrounding bedding when replacing the masked subject with a golden retriever puppy and recolors birds outside the mask. MaskFlow without SPD better localizes the edits, but visible discontinuities remain in the bedding texture and sky near the edited regions, as highlighted in the enlarged crops. The full model integrates the puppy more naturally with the bedding and recolors the selected bird while preserving the surrounding sky and other birds. These results show that SPD improves boundary continuity while maintaining regional control.

\textbf{Localization precision.}
We examine whether additional textual position descriptions are necessary when an explicit mask is provided. Both settings use editing instructions and masks, the proposed MaskFlow omits extra position descriptions during training, whereas the comparison model includes them. Table~\ref{tab:abl-precision} shows that omitting these descriptions yields consistent improvements across all reported metrics. This supports our training design, where the mask provides spatial guidance, allowing the instruction to focus on the desired edit. MaskFlow thus achieves effective regional control without requiring users to formulate detailed location descriptions.

\subsection{Sensitivity of MaskFlow Loss and Siamese Probability Paths}

\Figref{fig:abl-param} evaluates the loss weight $\omega_{\widetilde{\mM}}$ and background path exponent $\gamma$ on full images (Global) and foregrounds cropped using the masks (Crop).
As shown in~\Figref{fig:abl-param}, increasing $\omega_{\widetilde{\mM}}$ improves CLIP-I, particularly on foreground crops, with both evaluations peaking at $2.0$. A larger weight reduces performance, highlighting the importance of balancing regional and global supervision.
In SPP $\gamma$ settings, FID remains favorable up to $\gamma=1.5$ but deteriorates beyond it in both evaluations. This suggests that moderately faster background denoising provides useful context for foreground generation, while excessive separation between the paths weakens editing quality.
We therefore use $\omega_{\widetilde{\mM}}=2.0, \gamma=1.5$, the best setting in the comparisons.

\section{Conclusion}

This paper presents \textbf{MaskFlow}, a training and inference framework for regional image editing with precise localization, consistent background preservation, and seamless boundary transitions.
MaskFlow incorporates the mask into the flow-matching probability path and uses a mask-aware objective to improve localization accuracy and preserve unmasked content. Training on the prompts without additional position descriptions pushes the model to learn localization from masks, further strengthening the attentions to editable regions. The proposed Soft-Poisson De-seaming module refines the vector field during sampling to produce smooth transitions between edited and preserved regions. Moreover, the constructed MaskEdit-Benchmark provides a new regional editing dataset covering both general scene and infographics editing tasks for further study.
Quantitative and qualitative experiments demonstrate the effectiveness of MaskFlow on both natural scenes and infographic images, showing promise for practical application.

\clearpage
\bibliographystyle{iclr2027_conference}
\bibliography{iclr2027_conference}

\clearpage
\appendix
\section{Synthesized Data Analysis}
\label{apx:data-analysis}

\begin{figure}[htbp]
  \centering
  \begin{minipage}[c]{0.25\linewidth}
    \centering
    \includegraphics[width=\linewidth]{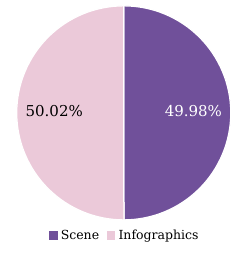}
  \end{minipage}\hfill
  \begin{minipage}[c]{0.25\linewidth}
    \centering
    \includegraphics[width=\linewidth]{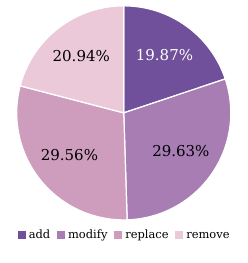}
  \end{minipage}\hfill
  \begin{minipage}[c]{0.48\linewidth}
    \centering
    \includegraphics[width=\linewidth]{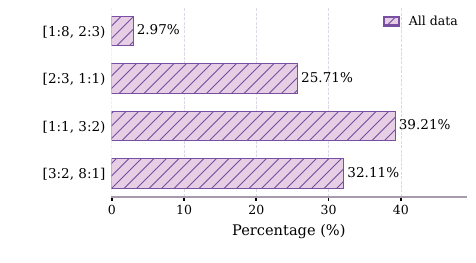}
  \end{minipage}
  \caption{Composition of MaskEdit-10k across all 11,213 paired samples. \textbf{Left:} source content subsets. \textbf{Middle:} four main regional image editing tasks. \textbf{Right:} source image aspect ratios.}
  \label{apx-fig:maskedit-data-statistics}
\end{figure}

We analyze MaskEdit-10k, the dataset produced by the synthesis pipeline in~\Secref{sec:data-pipeline}. It contains 11,213 editing samples, comprising 9,613 training samples and 1,600 test samples used for MaskEdit-Benchmark. Figure~\ref{apx-fig:maskedit-data-statistics} summarizes the complete dataset. Each editing example is counted once, and all percentages use the same total of 11,213 samples.

\textbf{Synthesis pipeline.}
The three stages in~\Secref{sec:data-pipeline} turn source images into localized editing examples. Object detection identifies salient editable objects, and prompt generation specifies the operation and desired result. Image generation produces the edited target, followed by SAM-based segmentation and human mask refinement.
Each example includes both a complete instruction with explicit location cues and a counterpart that refers to the masked region.

\textbf{Content source.}
The dataset contains 5,604 scene examples (49.98\%) and 5,609 infographics examples (50.02\%). The scene subset uses COCO-derived source images processed with super-resolution, whereas the infographics sources are synthetically generated.

\textbf{Editing operations.}
We derive task categories from the primary operations in the complete instructions and group them into \textit{add}, \textit{modify attributes}, \textit{replace objects}, and \textit{remove}. Their counts are 2,228 (19.87\%), 3,322 (29.63\%), 3,315 (29.56\%), and 2,348 (20.94\%), respectively. Modification includes changes to color, texture, and text content while preserving object shape, whereas replacement changes the edited object into another one.

\textbf{Aspect ratios.}
The dataset and benchmark cover both portrait and landscape source images. We compute the aspect ratio as $r=W/H$ from the stored source-image dimensions. The intervals $[1/8,2/3)$, $[2/3,1)$, $[1,3/2)$, and $[3/2,8]$ account for 2.97\%, 25.71\%, 39.21\%, and 32.11\% of all samples (333, 2,883, 4,397, and 3,600 examples), respectively. Intervals are left-closed and right-open, except for the final interval, which includes its right endpoint.

\section{Soft-Poisson De-seaming}
\label{apx:poisson-soft}

\subsection{Detailed formula derivation}

We follow the notation of \Secref{sec:preliminaries} and \Secref{sec:soft-poisson-deseaming}. Let $\vx_S$ denote the source image and $\vy$ a candidate image to be refined. In latent space, their feature maps have shape $C\times H\times W$, and $\vx_S,\vy\in\R^{CHW}$ denote their vectorized representations. For a spatial coordinate $\vp=(i,j)$, the corresponding feature vectors $\vx_S(\vp),\vy(\vp)$ belong to $\R^C$. The candidate is the edited target $\vy=\vx_0$ during training and the predicted clean sample $\vy=\widehat{\vx}_0^t$ at inference timestep $t$.

Let $\Omega$ denote the editable region and let $\mM\in\{0,1\}^{H\times W}$ be the binary mask. Its spatial representation is $m(\vp)=\mM_{i,j}=\1_{\vp\in\Omega}$.
Following the main text, we define the soft mask directly by Gaussian blur:
$\widetilde{\mM}=\operatorname{GaussianBlur}(\mM),
\widetilde m(\vp)=\widetilde{\mM}_{i,j}\in[0,1],
\widetilde{\Omega}=\left\{\vp=(i,j)\in\{1,\ldots,H\}\times\{1,\ldots,W\}\;\middle|\;\widetilde m(\vp)>0\right\}$.

We optimize the objective as in \Eqref{eq:spd} under the boundary constraint $\vz\vert_{\partial\widetilde{\Omega}}=\vx_S\vert_{\partial\widetilde{\Omega}}$
\begin{equation}
  \begin{aligned}
  \vz^{\star}=\argmin_{\vz}\;&
  \int_{\widetilde{\Omega}}
    \widetilde m(\vp)
    \left\lVert
      \nabla_{\vp}\vz(\vp)-\nabla_{\vp}\vy(\vp)
    \right\rVert_F^2
  \,\mathrm{d}\vp \\
  &+\int_{\widetilde{\Omega}}
    \widetilde m(\vp)
    \left\lVert\vz(\vp)-\vy(\vp)\right\rVert_2^2
  \,\mathrm{d}\vp \\
  &+\int_{\widetilde{\Omega}}
    \left(1-\widetilde m(\vp)\right)
    \left\lVert\vz(\vp)-\vx_S(\vp)\right\rVert_2^2
  \,\mathrm{d}\vp,
  \end{aligned}
  \label{app-eq:poisson-soft}
\end{equation}

Let $\mathcal{J}(\vz)$ denote the objective function
\begin{equation}
  \begin{aligned}
  \mathcal{J}(\vz)=\int_{\widetilde{\Omega}}\Big(&
    \widetilde m(\vp)\left\lVert
      \nabla_{\vp}\vz(\vp)-\nabla_{\vp}\vy(\vp)
    \right\rVert_F^2 \\
    &+\widetilde m(\vp)
    \left\lVert\vz(\vp)-\vy(\vp)\right\rVert_2^2 \\
    &+\left(1-\widetilde m(\vp)\right)
    \left\lVert\vz(\vp)-\vx_S(\vp)\right\rVert_2^2
  \Big)\,\mathrm{d}\vp.
  \end{aligned}
  \label{app-eq:poisson-energy}
\end{equation}
We solve this problem using a variational method. Consider a small perturbation $\alpha\bm{\eta}(\vp)$, where $\bm{\eta}(\vp)\in\R^C$ is an arbitrary differentiable vector field satisfying $\bm{\eta}\vert_{\partial\widetilde{\Omega}}=\vzero$. At an extremum, the first variation must satisfy
\begin{equation}
  \left.\frac{\mathrm{d}}{\mathrm{d}\alpha}
  \mathcal{J}(\vz+\alpha\bm{\eta})\right\vert_{\alpha=0}=0.
  \label{app-eq:soft-poisson-extremum}
\end{equation}

\textbf{Term 1.} Define the \textit{gradient-alignment} term as
\begin{equation}
  \mathcal{J}_{\mathrm g}(\vz+\alpha\bm{\eta})
  =\int_{\widetilde{\Omega}}
    \widetilde m(\vp)
    \left\lVert
      \nabla_{\vp}(\vz+\alpha\bm{\eta})
      -\nabla_{\vp}\vy
    \right\rVert_F^2
  \,\mathrm{d}\vp.
\end{equation}
Its derivative at $\alpha=0$ is

\begin{align}
  \left.\frac{\mathrm{d}}{\mathrm{d}\alpha}
  \mathcal{J}_{\mathrm g}(\vz+\alpha\bm{\eta})\right\vert_{\alpha=0}
  ={}&2\int_{\widetilde{\Omega}}\widetilde m(\vp)
  \left\langle\nabla_{\vp}\vz-\nabla_{\vp}\vy,
  \nabla_{\vp}\bm{\eta}\right\rangle_F\,\mathrm{d}\vp \\
  ={}&2\int_{\partial\widetilde{\Omega}}\widetilde m(\vp)
  \bm{\eta}^{\top}(\nabla_{\vp}\vz-\nabla_{\vp}\vy)\vn\,\mathrm{d}s \notag\\
  &-2\int_{\widetilde{\Omega}}\bm{\eta}^{\top}
  \nabla_{\vp}\cdot\left[\widetilde m(\vp)(\nabla_{\vp}\vz-\nabla_{\vp}\vy)\right]\,\mathrm{d}\vp \\
  ={}&-2\int_{\widetilde{\Omega}}\bm{\eta}^{\top}
  \nabla_{\vp}\cdot\left[\widetilde m(\vp)(\nabla_{\vp}\vz-\nabla_{\vp}\vy)\right]\,\mathrm{d}\vp,
  \label{app-eq:item1-eta}
\end{align}

where the second equality follows from integration by parts, $\vn$ is the outward unit normal vector, and the boundary integral vanishes because $\bm{\eta}\vert_{\partial\widetilde{\Omega}}=\vzero$.

\textbf{Term 2.} Define the \textit{edited-content} alignment term as
\begin{equation}
  \mathcal{J}_{\mathrm e}(\vz+\alpha\bm{\eta})
  =\int_{\widetilde{\Omega}}
  \widetilde m(\vp)
  \left\lVert
    (\vz+\alpha\bm{\eta})-\vy
  \right\rVert_2^2
  \,\mathrm{d}\vp.
\end{equation}
Its derivative at $\alpha=0$ is
\begin{equation}
  \left.\frac{\mathrm{d}}{\mathrm{d}\alpha}
  \mathcal{J}_{\mathrm e}(\vz+\alpha\bm{\eta})\right\vert_{\alpha=0}
  =2\int_{\widetilde{\Omega}}
  \widetilde m(\vp)
  \left(\vz-\vy\right)^{\top}\bm{\eta}
  \,\mathrm{d}\vp.
\end{equation}

\textbf{Term 3.} Define the \textit{source-alignment} term as
\begin{equation}
  \mathcal{J}_{\mathrm s}(\vz+\alpha\bm{\eta})
  =\int_{\widetilde{\Omega}}
  \left(1-\widetilde m(\vp)\right)
  \left\lVert
    (\vz+\alpha\bm{\eta})-\vx_S
  \right\rVert_2^2
  \,\mathrm{d}\vp.
\end{equation}
Its derivative at $\alpha=0$ is
\begin{equation}
  \left.\frac{\mathrm{d}}{\mathrm{d}\alpha}
  \mathcal{J}_{\mathrm s}(\vz+\alpha\bm{\eta})\right\vert_{\alpha=0}
  =2\int_{\widetilde{\Omega}}
  \left(1-\widetilde m(\vp)\right)
  \left(\vz-\vx_S\right)^{\top}\bm{\eta}
  \,\mathrm{d}\vp.
\end{equation}

Combining the three terms gives

\begin{equation}
  \begin{aligned}
  2\int_{\widetilde{\Omega}}\Big[&
  -\nabla_{\vp}\cdot\left[\widetilde m(\vp)(\nabla_{\vp}\vz-\nabla_{\vp}\vy)\right]
  +\widetilde m(\vp)(\vz-\vy)\\
  &+(1-\widetilde m(\vp))(\vz-\vx_S)
  \Big]^{\top}\bm{\eta}\,\mathrm{d}\vp=0.
  \end{aligned}
  \label{app-eq:first-variation}
\end{equation}

Because this equality holds for every admissible perturbation $\bm{\eta}$, the vector-valued integrand must vanish. Thus,

\begin{equation}
  \begin{aligned}
  \nabla_{\vp}\cdot\left[\widetilde m(\vp)(\nabla_{\vp}\vz(\vp)-\nabla_{\vp}\vy(\vp))\right]
  ={}&\widetilde m(\vp)(\vz(\vp)-\vy(\vp))\\
  &+(1-\widetilde m(\vp))(\vz(\vp)-\vx_S(\vp)).
  \end{aligned}
  \label{app-eq:poisson-euler}
\end{equation}

Since $\widetilde m+(1-\widetilde m)=1$, this equation can also be written as

\begin{equation}
  \begin{aligned}
  -\nabla_{\vp}\cdot\left[\widetilde m(\vp)\nabla_{\vp}\vz(\vp)\right]+\vz(\vp)
  ={}&-\nabla_{\vp}\cdot\left[\widetilde m(\vp)\nabla_{\vp}\vy(\vp)\right]\\
  &+\widetilde m(\vp)\vy(\vp)+(1-\widetilde m(\vp))\vx_S(\vp).
  \end{aligned}
  \label{app-eq:poisson-objective}
\end{equation}

with $\vz\vert_{\partial\widetilde{\Omega}}=\vx_S\vert_{\partial\widetilde{\Omega}}$. The unit zeroth-order term makes this a weighted screened Poisson equation. This equation will be derived to~\Eqref{eq:spd-jacobi} in the following sections.

\subsection{Discretization}

To solve \Eqref{app-eq:poisson-objective} in the discrete latent feature space with efficient numerical methods, we discretize the spatial differential operators on a grid with unit spacing. Let $\vp=(i,j)$ be a pixel in $\widetilde{\Omega}$, let $\operatorname{N}(\vp)$ denote its neighboring pixels, and let $\vz_{\vp}\in\R^C$ be the feature vector at $\vp$. Define $w_{\vp\vq}=(\widetilde m_{\vp}+\widetilde m_{\vq})/2$ and $J_{\vp}=\sum_{\vq\in\operatorname{N}(\vp)}w_{\vp\vq}$. We discretize $-\nabla_{\vp}\cdot(\widetilde m\nabla_{\vp})$ using the arithmetic mean of the soft-mask values at adjacent grid points. The resulting weighted negative Laplacian $\mL$ is
$(\mL\vz)_{\vp}
  =\sum_{\vq\in\operatorname{N}(\vp)}w_{\vp\vq}
    (\vz_{\vp}-\vz_{\vq})
  =J_{\vp}\vz_{\vp}
  -\sum_{\vq\in\operatorname{N}(\vp)}w_{\vp\vq}\vz_{\vq}$,
where the boundary condition $\vz\vert_{\partial\widetilde{\Omega}}=\vx_S\vert_{\partial\widetilde{\Omega}}$ remains in effect. For a neighbor $\vq\in\operatorname{N}(\vp)\cap\widetilde{\Omega}$, the value $\vz_{\vq}$ is unknown and must be solved. For a neighbor $\vq\in\operatorname{N}(\vp)\backslash\widetilde{\Omega}$, the boundary condition provides the known value $\vz_{\vq}=\vx_{S,\vq}$.
Therefore,
\begin{equation}
  (\mL\vz)_{\vp}
  =J_{\vp}\vz_{\vp}
  -\sum_{\vq\in\operatorname{N}(\vp)\cap\widetilde{\Omega}}w_{\vp\vq}\vz_{\vq}
  -\sum_{\vq\in\operatorname{N}(\vp)\backslash\widetilde{\Omega}}w_{\vp\vq}\vx_{S,\vq}.
  \label{app-eq:discrete-z_p}
\end{equation}

Using \Eqref{app-eq:discrete-z_p}, we discretize \Eqref{app-eq:poisson-objective}. Let $\widetilde m_{\vp}$ be the soft-mask value at $\vp$. The diagonal coefficient is
  $J_{\vp}+\widetilde m_{\vp}+\left(1-\widetilde m_{\vp}\right)=J_{\vp}+1$.
The discrete system is
\begin{equation}
  \resizebox{0.90\linewidth}{!}{$\displaystyle
  (J_{\vp}+1)\vz_{\vp}
  -\sum_{\vq\in\operatorname{N}(\vp)\cap\widetilde{\Omega}}w_{\vp\vq}\vz_{\vq}
  =
  \sum_{\vq\in\operatorname{N}(\vp)\backslash\widetilde{\Omega}}w_{\vp\vq}\vx_{S,\vq}
  +\sum_{\vq\in\operatorname{N}(\vp)}w_{\vp\vq}
    \left(\vy_{\vp}-\vy_{\vq}\right)
  +\widetilde m_{\vp}\vy_{\vp}
  +(1-\widetilde m_{\vp})\vx_{S,\vp}.
  $}
  \label{app-eq:poisson-discrete-objective}
\end{equation}
For an interior grid point, the four-neighbor stencil is
\begin{equation}
  \operatorname{N}(\vp)
  =\{(i+1,j),(i-1,j),(i,j+1),(i,j-1)\},
  \qquad
  \left\lvert\operatorname{N}(\vp)\right\rvert=4.
\end{equation}
At image edges, we include only valid neighbors in $\operatorname{N}(\vp)$ and adjust $J_{\vp}$ accordingly. The weighted neighborhood sums can be implemented efficiently using convolution.

\subsection{Jacobi iteration}

We approximate the solution of \Eqref{app-eq:poisson-discrete-objective} using Jacobi iteration. Starting from $\vz_{\vp}^{(0)}=\vy_{\vp}$ inside $\widetilde{\Omega}$ and fixing the values outside it to the source, the update is
\begin{equation}
  \resizebox{0.90\linewidth}{!}{$\displaystyle
  \vz_{\vp}^{(k+1)}=\frac{1}{J_{\vp}+1}\Bigg[
  \sum_{\vq\in\operatorname{N}(\vp)\cap\widetilde{\Omega}}w_{\vp\vq}\vz_{\vq}^{(k)}
  +\sum_{\vq\in\operatorname{N}(\vp)\backslash\widetilde{\Omega}}w_{\vp\vq}\vx_{S,\vq}
  +\sum_{\vq\in\operatorname{N}(\vp)}w_{\vp\vq}
    \left(\vy_{\vp}-\vy_{\vq}\right)
  +\widetilde m_{\vp}\vy_{\vp}
  +(1-\widetilde m_{\vp})\vx_{S,\vp}
  \Bigg].
  $}
  \label{app-eq:poisson-jacobi}
\end{equation}
This is the update in \Eqref{eq:spd-jacobi}. The system is strictly diagonally dominant because $J_{\vp}+1>\sum_{\vq\in\operatorname{N}(\vp)\cap\widetilde{\Omega}}w_{\vp\vq}$, ensuring convergence of Jacobi iteration. After $K\in\sN^+$ iterations, we form the refined candidate
\begin{equation}
  \vy_{\vp}^{\star}\approx
  \begin{cases}
    \vz_{\vp}^{(K)}, & \vp\in\widetilde{\Omega},\\
    \vx_{S,\vp}, & \vp\notin\widetilde{\Omega}.
  \end{cases}
  \label{app-eq:refined-candidate}
\end{equation}

\subsection{Vector field refinement with Soft-Poisson De-seaming}
\label{apx:clean-sample-prediction}

This section derives the bias item $C(\sigma_t, \gamma)$ at inference. For the siamese path in Equations~\ref{eq:spp} and~\ref{eq:spp-gamma}, define the clean endpoint as $\widehat{\vx}_{0}^t=\widetilde m\odot\vx_0^{\star}+(1-\widetilde m)\odot\vx_S$. The target vector field is
\begin{equation}
  \vv_{\sigma}
  =\widetilde m\odot(\epsilon-\vx_0^{\star})
  +(1-\widetilde m)\odot\gamma\sigma^{\gamma-1}(\epsilon-\vx_S).
  \label{app-eq:bias-velocity}
\end{equation}
Subtracting $\sigma\vv_{\sigma}$ from $\vx_{\sigma}$ cancels the linear foreground displacement and leaves
\begin{equation}
  \vx_{\sigma}-\sigma\vv_{\sigma}
  =\widehat{\vx}_{0}^t
  +(1-\widetilde m)\odot(1-\gamma)\sigma^{\gamma}(\epsilon-\vx_S).
  \label{app-eq:bias-residual}
\end{equation}
Thus, $\widehat{\vx}_{0}^t=\vx_{\sigma}-\sigma\vv_{\sigma}+C(\sigma,\gamma)$, where
$C(\sigma,\gamma)
  =(1-\widetilde m)\odot(\gamma-1)\sigma^{\gamma}(\epsilon-\vx_S)$.

During sampling, we use the learned velocity to estimate the clean endpoint at $\sigma_t>0$:
\begin{equation}
  \widehat{\vx}_0^t
  =\vx_{\sigma_t}
  -\sigma_t\vv_{\vtheta}(\vx_{\sigma_t},\sigma_t,\vc)
  +C(\sigma_t,\gamma).
  \label{app-eq:target-estimate}
\end{equation}
Applying SPD to $\vy=\widehat{\vx}_0^t$ yields $\widehat{\vx}_0^{t,\star}$. The refined velocity is
\begin{equation}
  \vv_{\vtheta}^{\star}(\vx_{\sigma_t},\sigma_t,\vc)
  =\frac{\vx_{\sigma_t}+C(\sigma_t,\gamma)-\widehat{\vx}_0^{t,\star}}{\sigma_t}
  =\vv_{\vtheta}(\vx_{\sigma_t},\sigma_t,\vc)
    +\frac{\widehat{\vx}_0^t-\widehat{\vx}_0^{t,\star}}{\sigma_t}.
  \label{app-eq:refined-vector-field}
\end{equation}
The Euler update in $\sigma$ is then
$\vx_{\sigma_{t-1}}
  =\vx_{\sigma_t}
  +(\sigma_{t-1}-\sigma_t)
  \vv_{\vtheta}^{\star}(\vx_{\sigma_t},\sigma_t,\vc)$.
We repeat these steps for $t=T,\ldots,1$ with $1=\sigma_T>\cdots>\sigma_0=0$, starting from $\vx_{\sigma_T}=\epsilon$.

\section{Training and Inference Algorithms}
\label{apx:algrithms}

We demonstrate the whole training and inference workflow in this section to clearly describe our proposed MaskFlow. The framework follows the flow-matching, but with novel mask-aware modifications. Following the main text, $\mM$ denotes the binary input mask, $\widetilde{\mM}=\operatorname{GaussianBlur}(\mM)$ is the soft mask, and $\widetilde m(\vp)=\widetilde{\mM}_{i,j}$ denotes its spatial weights, broadcast across feature channels for element-wise operations.

\begin{figure}[t]
\centering
\begin{minipage}[t]{0.49\textwidth}
\vspace{0pt}
\begin{algorithm}[H]
\caption{Training}
\label{apx-alg:maskflow-training}
\footnotesize
\begin{algorithmic}[1]
\Require Dataset $\mathcal{D}$, predictor $v_\theta$,
         schedule $\sigma(\cdot)$, exponent $\gamma>1$
\Ensure Updated parameters $\theta$

\State Sample $(x_0,x_S,c,\mM)\sim\mathcal{D}$
\State $\widetilde{\mM}\gets\operatorname{GaussianBlur}(\mM)$
\State Sample $t\sim\mathcal{U}(0,1)$; $\sigma_t\gets\sigma(t)$; $\epsilon\sim\mathcal{N}(0,I)$

\State $x_0^\star\gets\operatorname{SPD}(x_0,x_S,\widetilde{\mM})$
       \Comment{\Eqref{eq:spd-jacobi}}
\State $a\gets\epsilon-x_0^\star$; $b\gets\epsilon-x_S$

\Statex \textit{Construct noisy sample (\Eqref{eq:spp}):}
\State $\begin{aligned}
    x_t\gets{}&\widetilde m\odot(x_0^\star+\sigma_t a) \\
    &+(1-\widetilde m)\odot(x_S+\sigma_t^\gamma b)
\end{aligned}$

\Statex \textit{Compute target velocity:}
\State $\begin{aligned}
    u_t\gets{}\widetilde m\odot a+(1-\widetilde m)\odot\gamma\sigma_t^{\gamma-1}b
\end{aligned}$

\State $v\gets v_\theta(x_t,x_S,\widetilde{\mM},t,c)$
\State $\mathcal{L}_{\mathrm{FM}}
       \gets\operatorname{mean}((v-u_t)^2)$
\State $\begin{aligned}
    \mathcal{L}_{\mathrm{MaskFlow}}\gets{}&
    \frac{HW}{\sum_{H,W}\widetilde{\mM}}\\
    &\cdot\operatorname{mean}\!\left(\widetilde m\odot(v-u_t)^2\right)
\end{aligned}$

\State $\mathcal{L}\gets
       \mathcal{L}_{\mathrm{FM}}+\omega_{\widetilde{\mM}}\mathcal{L}_{\mathrm{MaskFlow}}$
\State Update $\theta$ using $\nabla_\theta\mathcal{L}$
\end{algorithmic}
\end{algorithm}
\end{minipage}\hfill
\begin{minipage}[t]{0.49\textwidth}
\vspace{0pt}
\begin{algorithm}[H]
\caption{Inference}
\label{apx-alg:maskflow-inference}
\footnotesize
\begin{algorithmic}[1]
\Require Source $x_S$, binary mask $\mM$, prompt $c$,
         predictor $v_\theta$, exponent $\gamma>1$
\Ensure Edited image $x_0$

\State $\widetilde{\mM}\gets\operatorname{GaussianBlur}(\mM)$
\State Choose $T\geq2$ noise levels:
\Statex \hspace{\algorithmicindent}
        $1=\sigma_{T-1}>\cdots>\sigma_0=0$
\State Sample $\epsilon\sim\mathcal{N}(0,I)$
\State $x_{T-1}\gets\epsilon$; $b\gets\epsilon-x_S$

\For{$t=T-1,T-2,\ldots,1$}
    \State $v\gets v_\theta(x_t,x_S,\widetilde{\mM},t,c)$

    \Statex \hspace{\algorithmicindent}
            \textit{Predict clean sample:}
    \State $\begin{aligned}
        \widehat{x}_0\gets{}&x_t-\sigma_t v \\
        &+(1-\widetilde m)\odot(\gamma-1)\sigma_t^\gamma b
    \end{aligned}$

    \State $\widehat{x}_0^\star
           \gets\operatorname{SPD}(\widehat{x}_0,x_S,\widetilde{\mM})$

    \Statex \hspace{\algorithmicindent}
            \textit{Refine velocity:}
    \State $v^\star\gets
           v+(\widehat{x}_0-\widehat{x}_0^\star)/\sigma_t$

    \Statex \hspace{\algorithmicindent}
            \textit{Denoising step:}
    \State $x_{t-1}\gets
           x_t+(\sigma_{t-1}-\sigma_t)v^\star$
\EndFor

\State \Return $x_0$
\end{algorithmic}
\end{algorithm}
\end{minipage}
\caption{The training and inference algorithms of the proposed MaskFlow framework.}
\end{figure}

\paragraph{Training.} As shown in~\Algref{apx-alg:maskflow-training}, given the prepared dataset $\mathcal{D}$, the vector predictor $\vv_{\vtheta}$ to be trained, and the noise schedule generator $\sigma(\cdot)$ with time-shift, the goal is to update the parameters in predictor to generate refined velocities on the whole source image based on the user editing purposes and freestyle mask regions.
The training start by sampling data tuple from dataset and noise from $\mathcal{N}(0,I)$, different from naive flow-matching which add sigma-modulated noise to the target feature, MaskFlow firstly blends the source background and the target foreground with Soft-Poisson De-seaming using~\Eqref{eq:spd-jacobi}. With the proposed Siamese Probability Paths, the latent feature at the sampled timestep $t$ is synthesized. Then, the conditional velocities in the mask and background are constructed. The predictor generate a velocity conditioned by the source and mask images, which is used for flow-matching loss calculation and the proposed MaskFlow loss for parameters updating.

\paragraph{Inference.} The inference pipeline starts at a pure Gaussian noise, the given source image and the user defined prompt and mask. The denoising process is iterative by traversing the timesteps. At each timestep, the trained vector predictor produces a vector. Then the clean sample is calculated for Soft-Poisson De-seaming refinement, which is different from traditional flow-matching, an additional bias should be added to the clean sample. Based on the re-generated clean sample, the refined velocity is updated by the differences. And finally, the next latent feature is updated by the refined velocity, ready for next iteration.

\Needspace{20\baselineskip}
\section{Soft-Poisson De-seaming Efficiency}
\label{apx:experiment}

\begin{wrapfigure}{r}{0.55\textwidth}
   \vspace{-\intextsep}
   \centering
   \includegraphics[width=\linewidth]{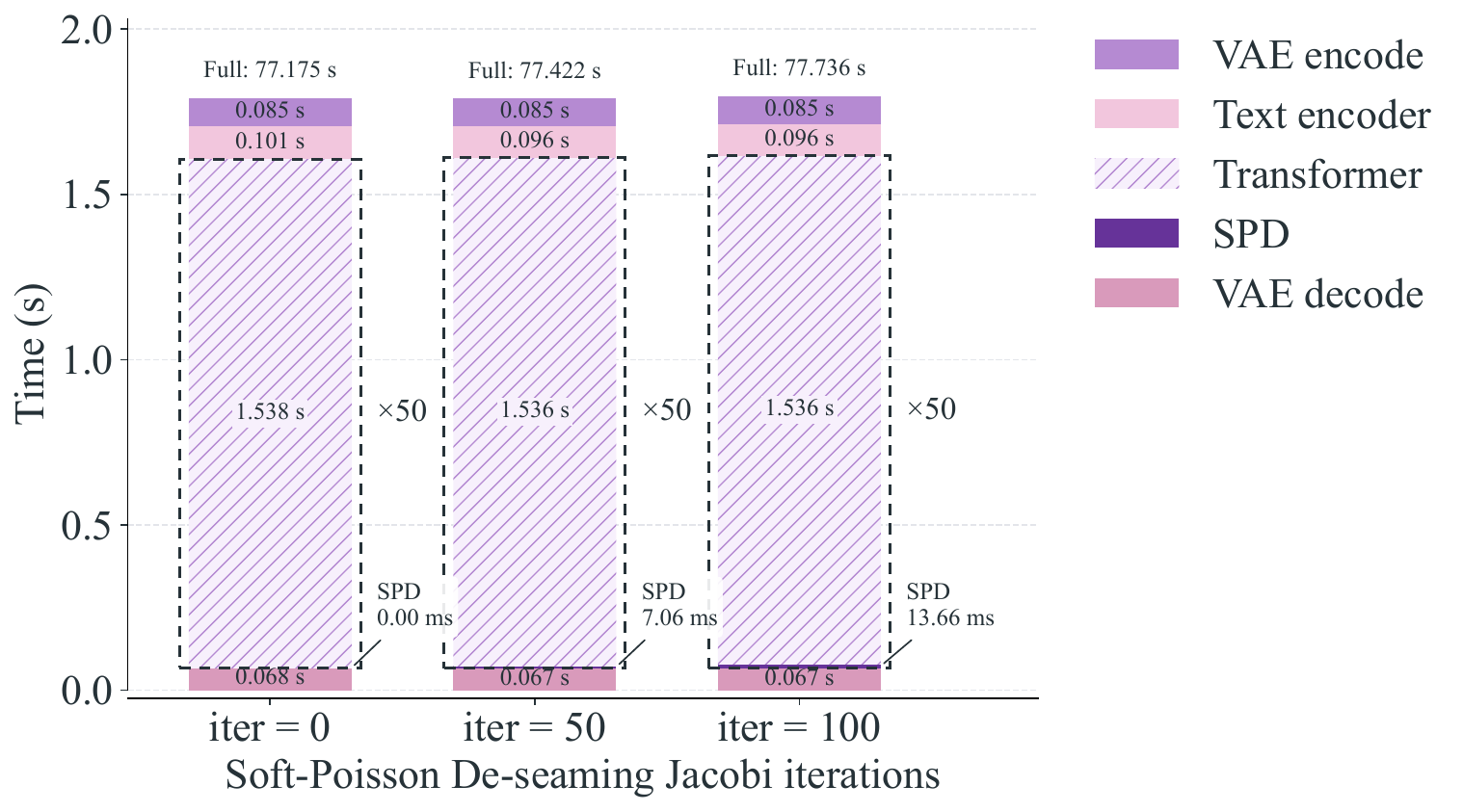}
   \vspace{-\intextsep}
   \caption{The latency of Soft-Poisson De-seaming module with different Jacobi iterations.}
   \label{apx-fig:spd-latency}
\end{wrapfigure}

We report the latency of the proposed Soft-Poisson De-seaming solved by Jacobi iterations. To evaluate the stable latency at inference time, we conduct $2$ additional warmup complete loops and start timing after warmup. As shown in~\Figref{apx-fig:spd-latency}, the latency of each module is the average of times spent by $10$ samples inference with $50$ steps. As it illustrated, the SPD module refines the vector field with a tiny contribution to overall inference. In practice, Soft-Poisson De-seaming solve the equation with maximum $50$ steps, where the total latency contribution can be ignored, showing the efficiency of the proposed method.

\section{More Visualization}
\label{apx:visualization}

This section presents additional qualitative comparisons between MaskFlow and ten baseline methods on scene and infographic image editing. For each dataset, we show four examples for each of four tasks: addition, modification, replacement, and removal. In each comparison, the first panel shows the source image with the editable region highlighted in translucent reddish purple; the remaining panels show the outputs of the compared methods. MaskFlow appears in the final panel, and the editing prompt is displayed below each comparison.

\newsavebox{\maskflowtaskbox}
\newlength{\maskflowtaskheight}

\clearpage

\subsection{Scene Image Editing}
\label{apx:visualization-scene}

\subsubsection{Addition}

\setlength{\maskflowtaskheight}{\dimexpr\textheight-\pagetotal-55pt\relax}
\begin{figure}[!ht]
  \centering
  \begin{lrbox}{\maskflowtaskbox}
    \begin{minipage}{\linewidth}
      \centering
      \subfloat{%
        \includegraphics[width=\linewidth]{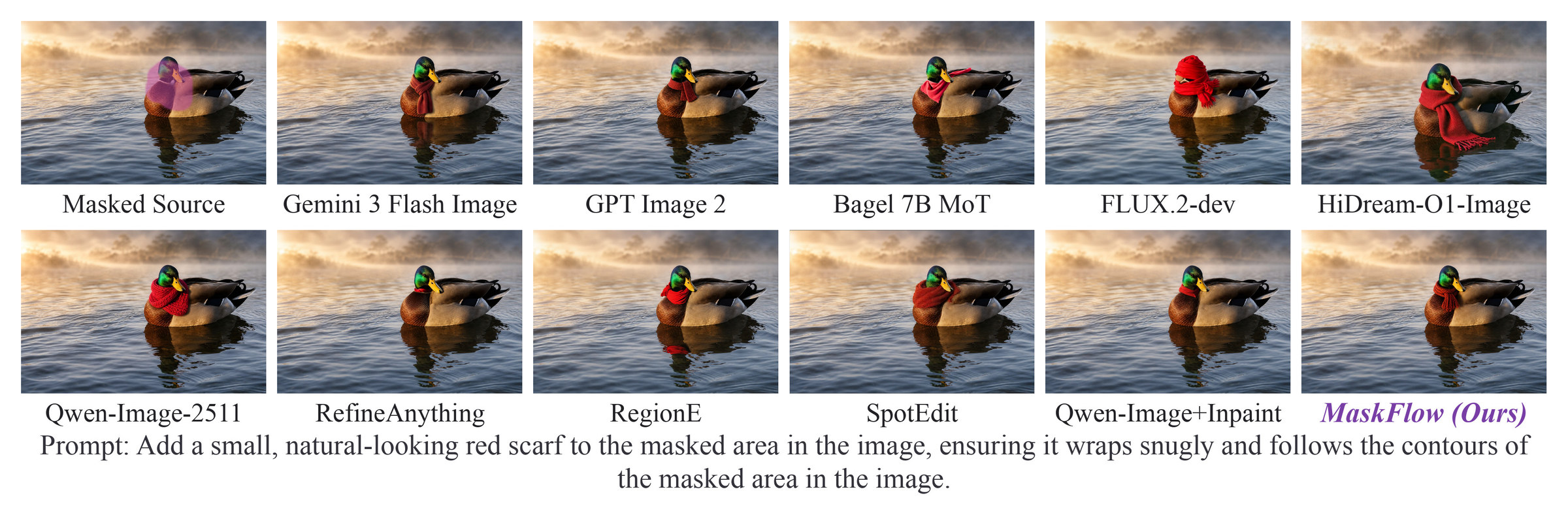}%
      }\par\vspace{2pt}
      \subfloat{%
        \includegraphics[width=\linewidth]{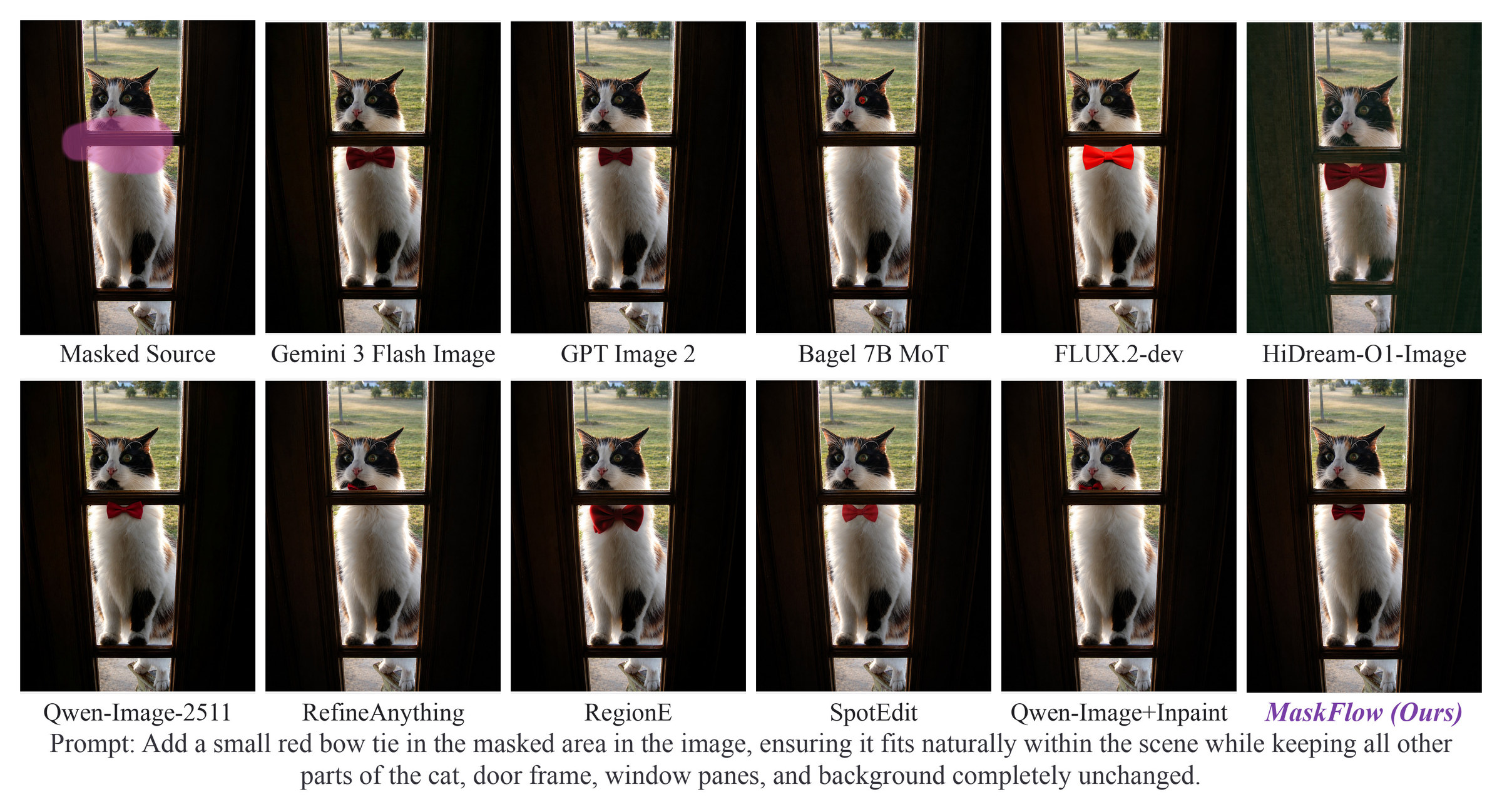}%
      }\par\vspace{2pt}
      \subfloat{%
        \includegraphics[width=\linewidth]{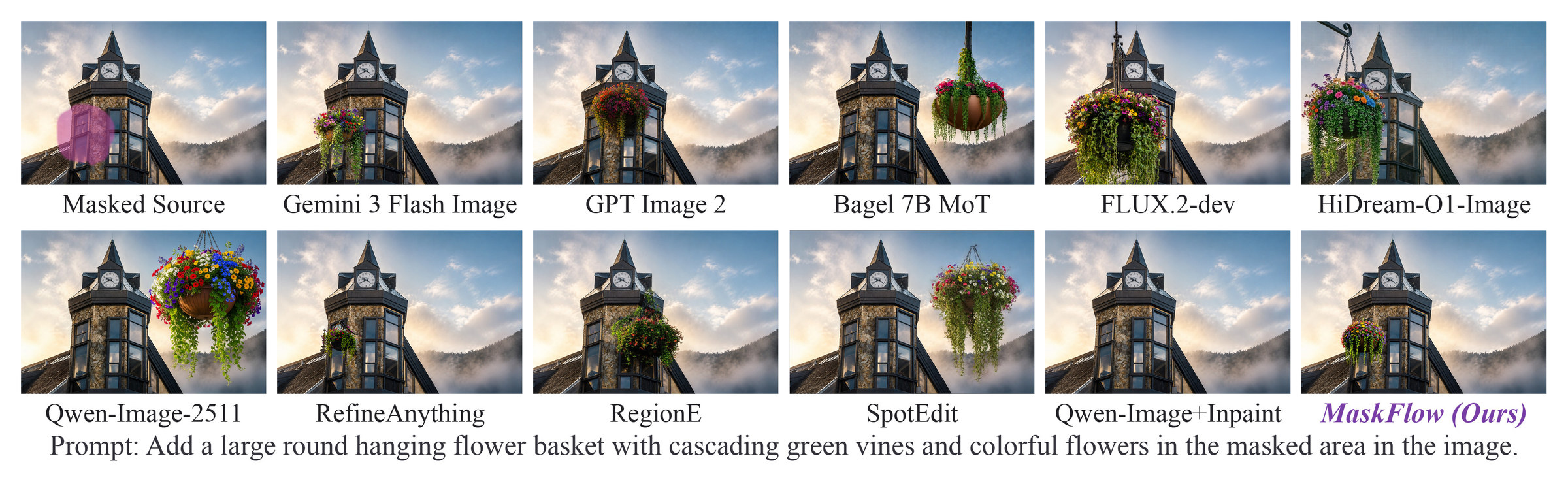}%
      }\par\vspace{2pt}
      \subfloat{%
        \includegraphics[width=\linewidth]{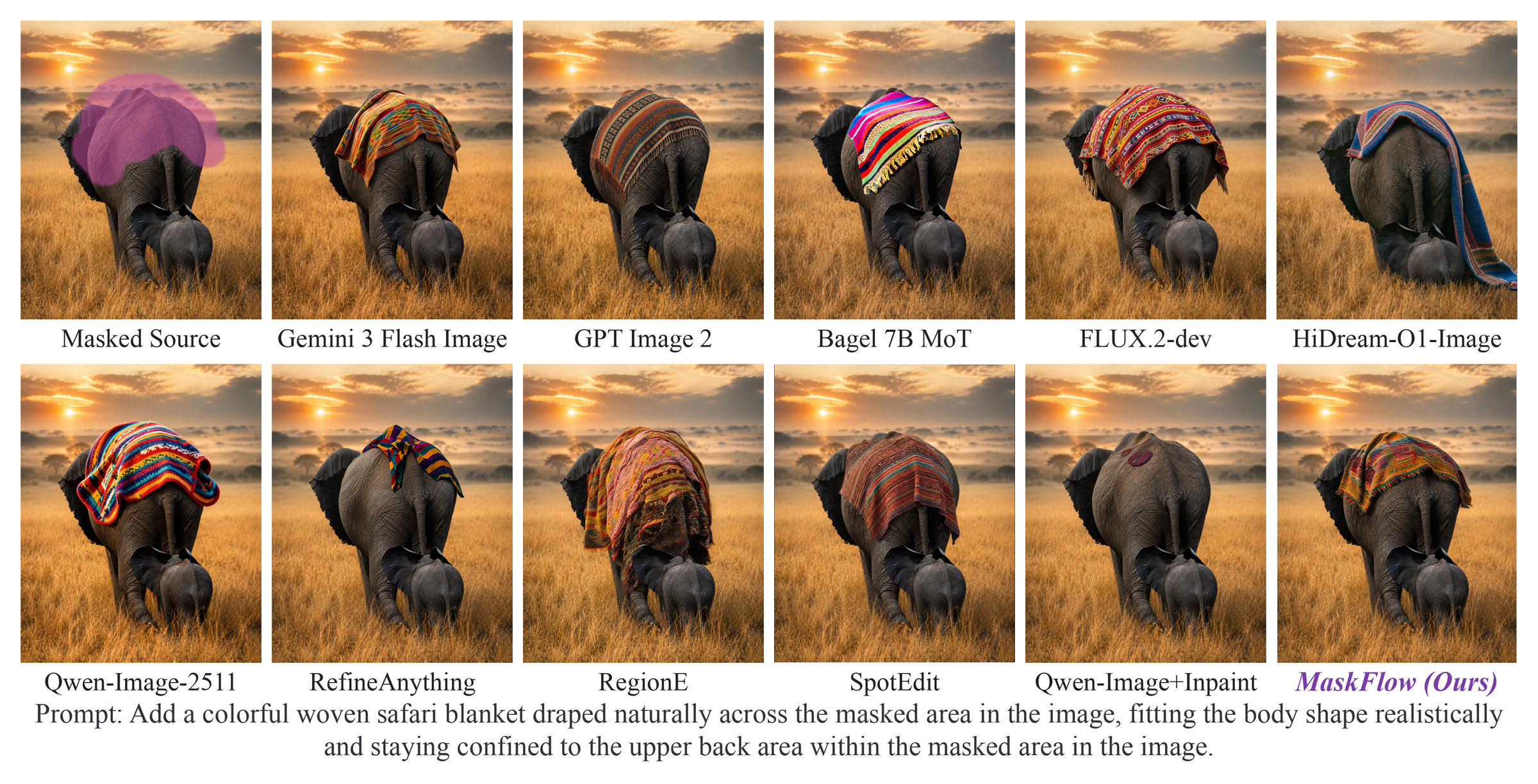}%
      }
    \end{minipage}
  \end{lrbox}
  \ifdim\dimexpr\ht\maskflowtaskbox+\dp\maskflowtaskbox\relax>\maskflowtaskheight
    \resizebox*{!}{\maskflowtaskheight}{\usebox{\maskflowtaskbox}}%
  \else
    \usebox{\maskflowtaskbox}%
  \fi
  \caption{Qualitative comparisons for addition on scene images.}
  \label{apx-fig:scene-add}
\end{figure}

\clearpage

\subsubsection{Modification}

\setlength{\maskflowtaskheight}{\dimexpr\textheight-\pagetotal-55pt\relax}
\begin{figure}[!ht]
  \centering
  \begin{lrbox}{\maskflowtaskbox}
    \begin{minipage}{\linewidth}
      \centering
      \subfloat{%
        \includegraphics[width=\linewidth]{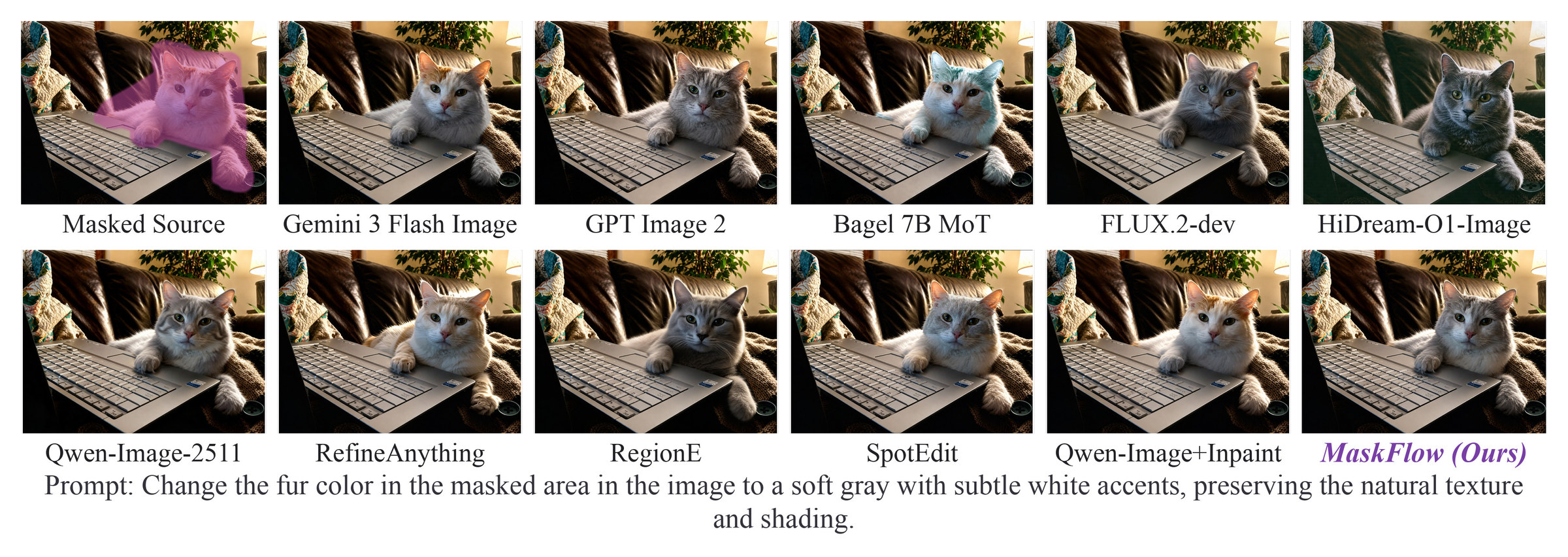}%
      }\par\vspace{2pt}
      \subfloat{%
        \includegraphics[width=\linewidth]{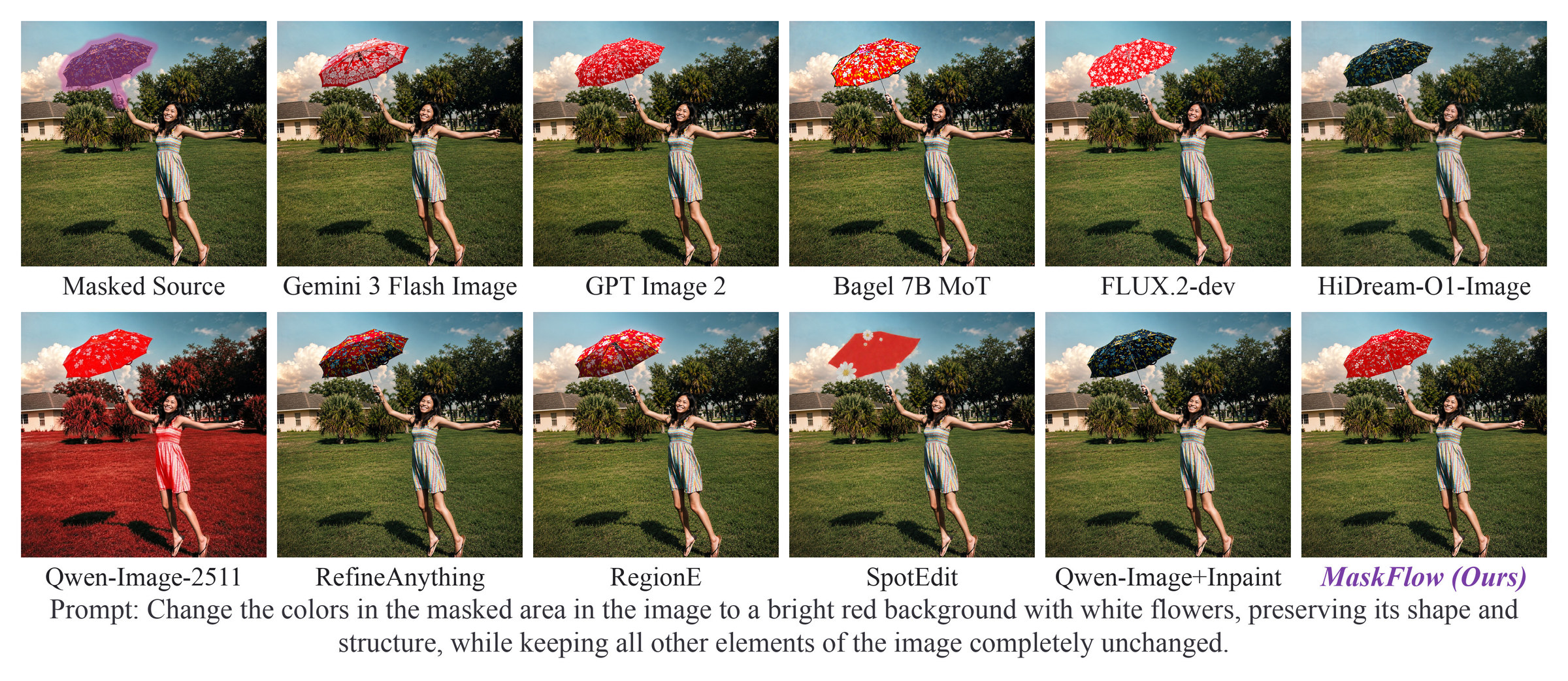}%
      }\par\vspace{2pt}
      \subfloat{%
        \includegraphics[width=\linewidth]{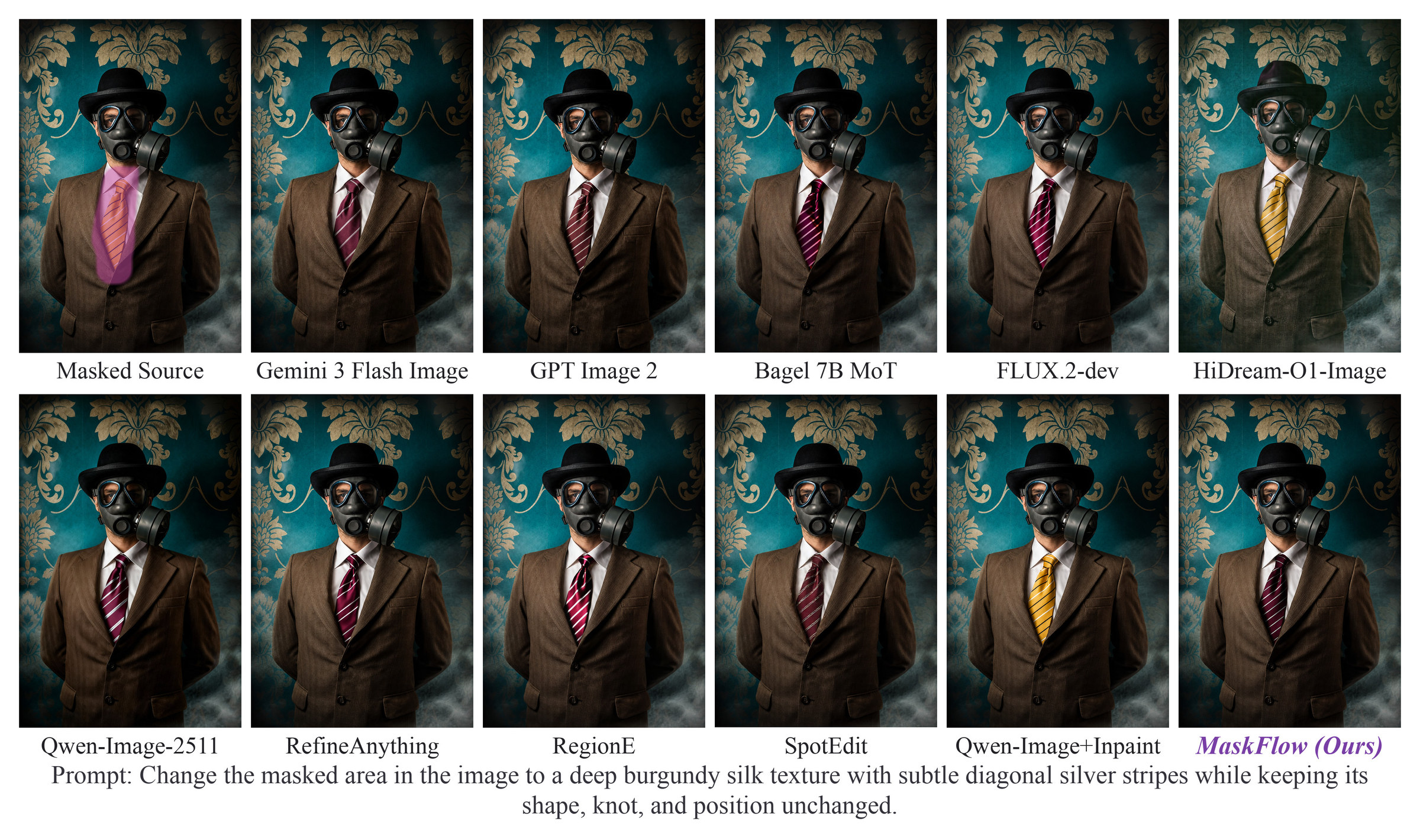}%
      }\par\vspace{2pt}
      \subfloat{%
        \includegraphics[width=\linewidth]{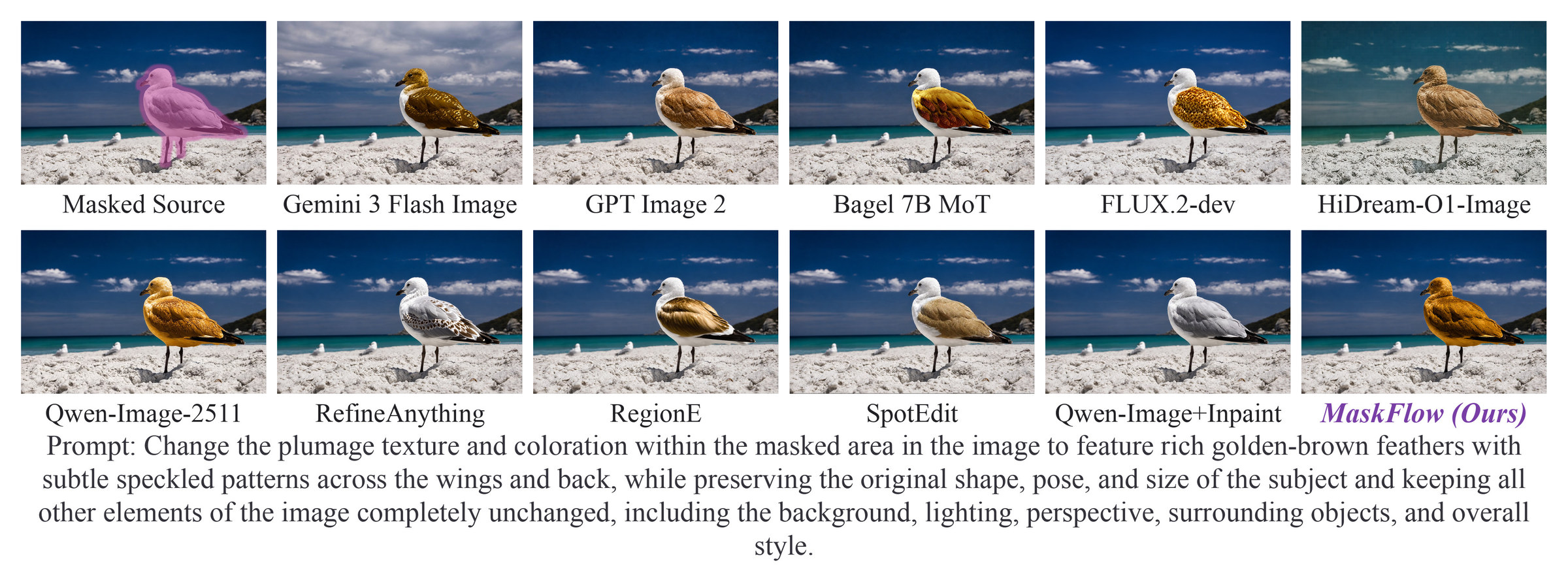}%
      }
    \end{minipage}
  \end{lrbox}
  \ifdim\dimexpr\ht\maskflowtaskbox+\dp\maskflowtaskbox\relax>\maskflowtaskheight
    \resizebox*{!}{\maskflowtaskheight}{\usebox{\maskflowtaskbox}}%
  \else
    \usebox{\maskflowtaskbox}%
  \fi
  \caption{Qualitative comparisons for modification on scene images.}
  \label{apx-fig:scene-modify}
\end{figure}

\clearpage

\subsubsection{Replacement}

\setlength{\maskflowtaskheight}{\dimexpr\textheight-\pagetotal-55pt\relax}
\begin{figure}[!ht]
  \centering
  \begin{lrbox}{\maskflowtaskbox}
    \begin{minipage}{\linewidth}
      \centering
      \subfloat{%
        \includegraphics[width=\linewidth]{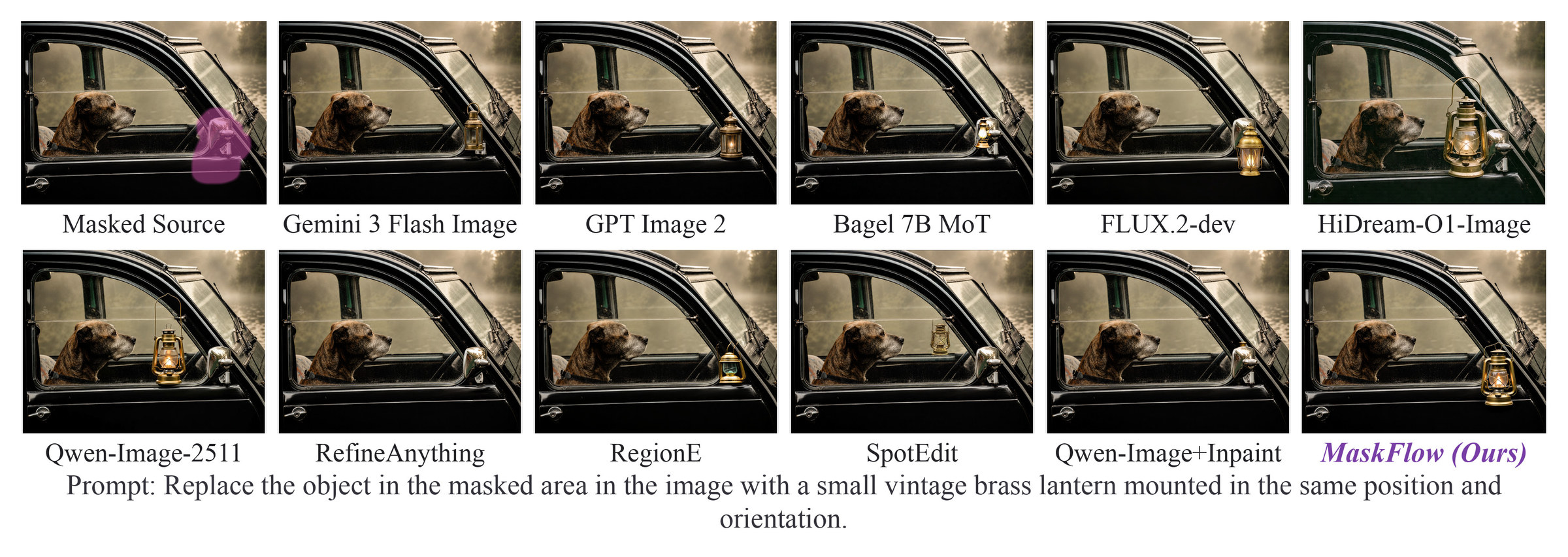}%
      }\par\vspace{2pt}
      \subfloat{%
        \includegraphics[width=\linewidth]{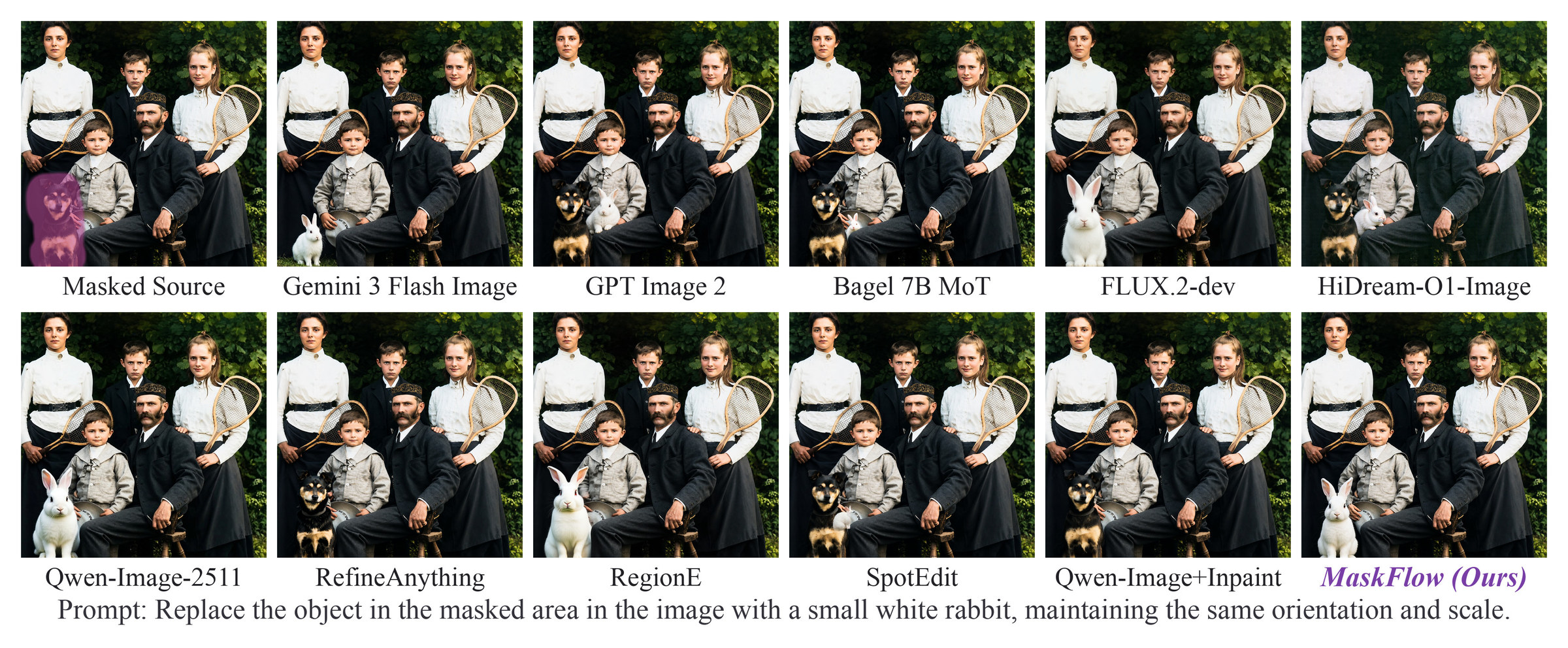}%
      }\par\vspace{2pt}
      \subfloat{%
        \includegraphics[width=\linewidth]{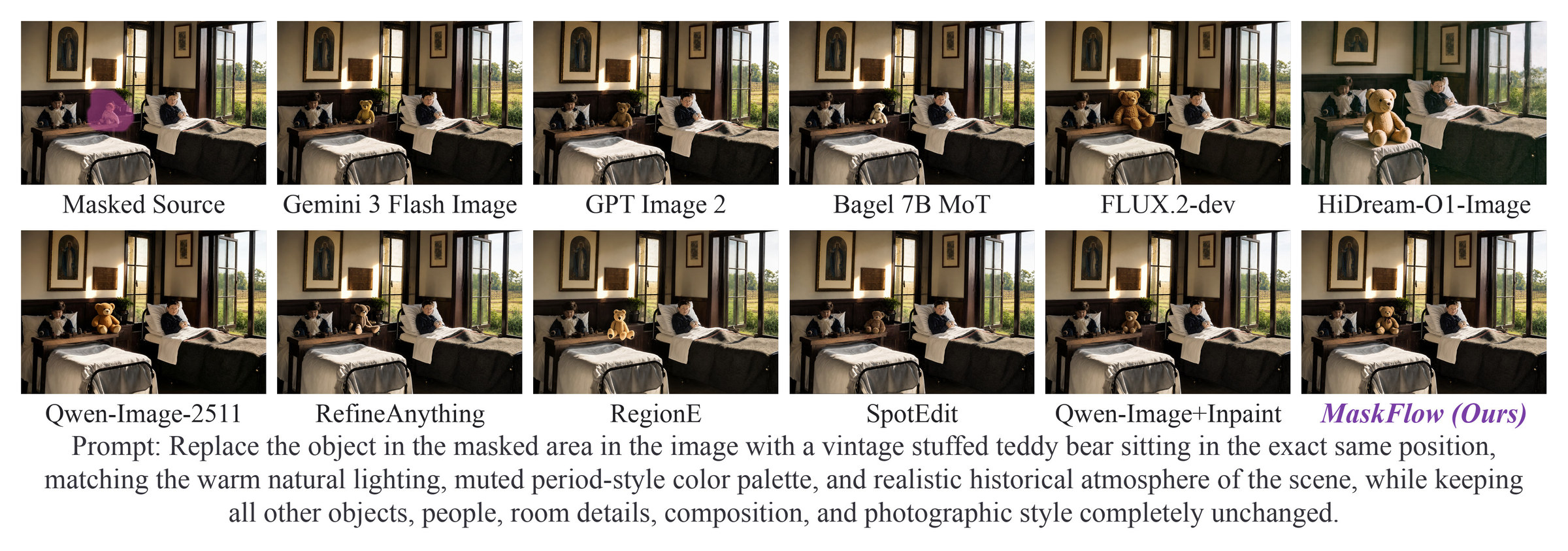}%
      }\par\vspace{2pt}
      \subfloat{%
        \includegraphics[width=\linewidth]{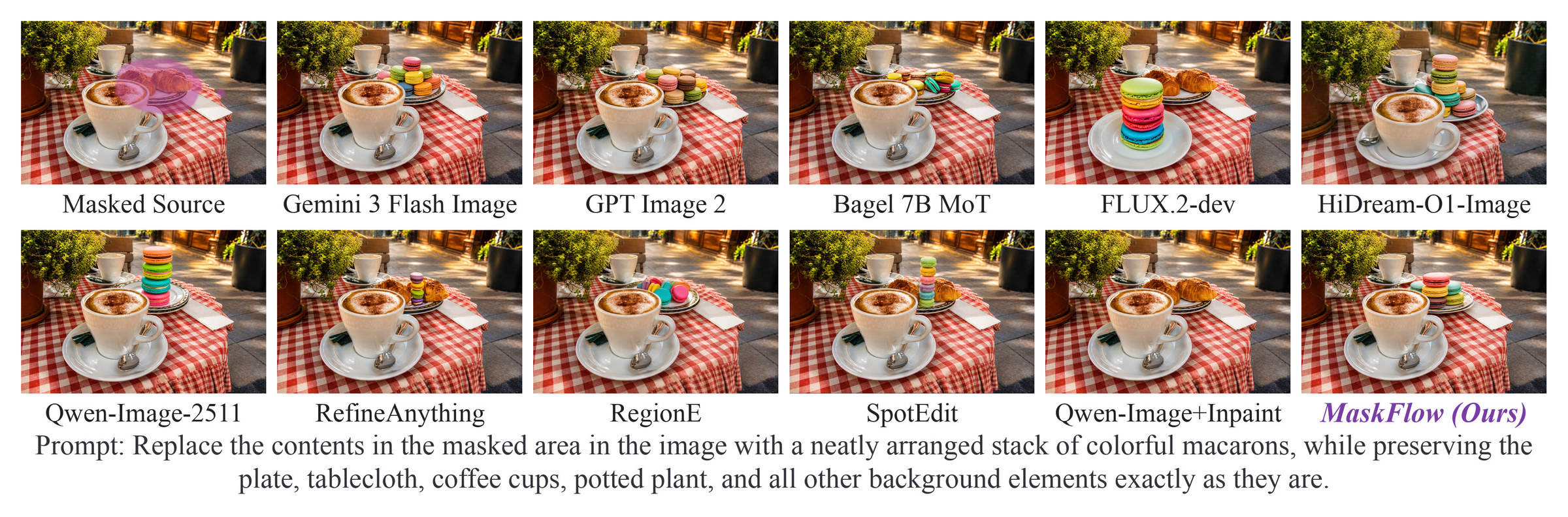}%
      }
    \end{minipage}
  \end{lrbox}
  \ifdim\dimexpr\ht\maskflowtaskbox+\dp\maskflowtaskbox\relax>\maskflowtaskheight
    \resizebox*{!}{\maskflowtaskheight}{\usebox{\maskflowtaskbox}}%
  \else
    \usebox{\maskflowtaskbox}%
  \fi
  \caption{Qualitative comparisons for replacement on scene images. }
  \label{apx-fig:scene-replace}
\end{figure}

\clearpage

\subsubsection{Removal}

\setlength{\maskflowtaskheight}{\dimexpr\textheight-\pagetotal-55pt\relax}
\begin{figure}[!ht]
  \centering
  \begin{lrbox}{\maskflowtaskbox}
    \begin{minipage}{\linewidth}
      \centering
      \subfloat{%
        \includegraphics[width=\linewidth]{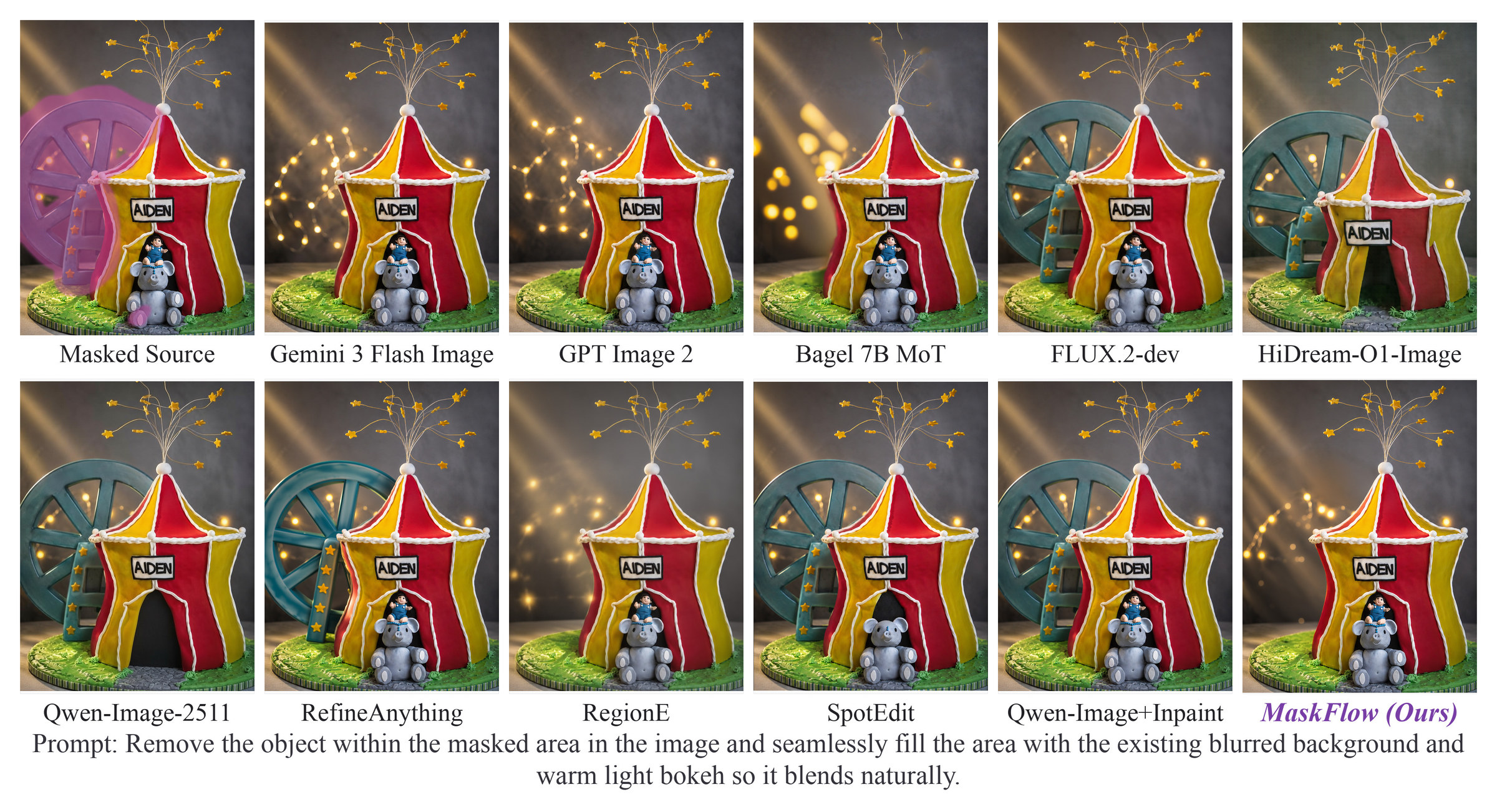}%
      }\par\vspace{2pt}
      \subfloat{%
        \includegraphics[width=\linewidth]{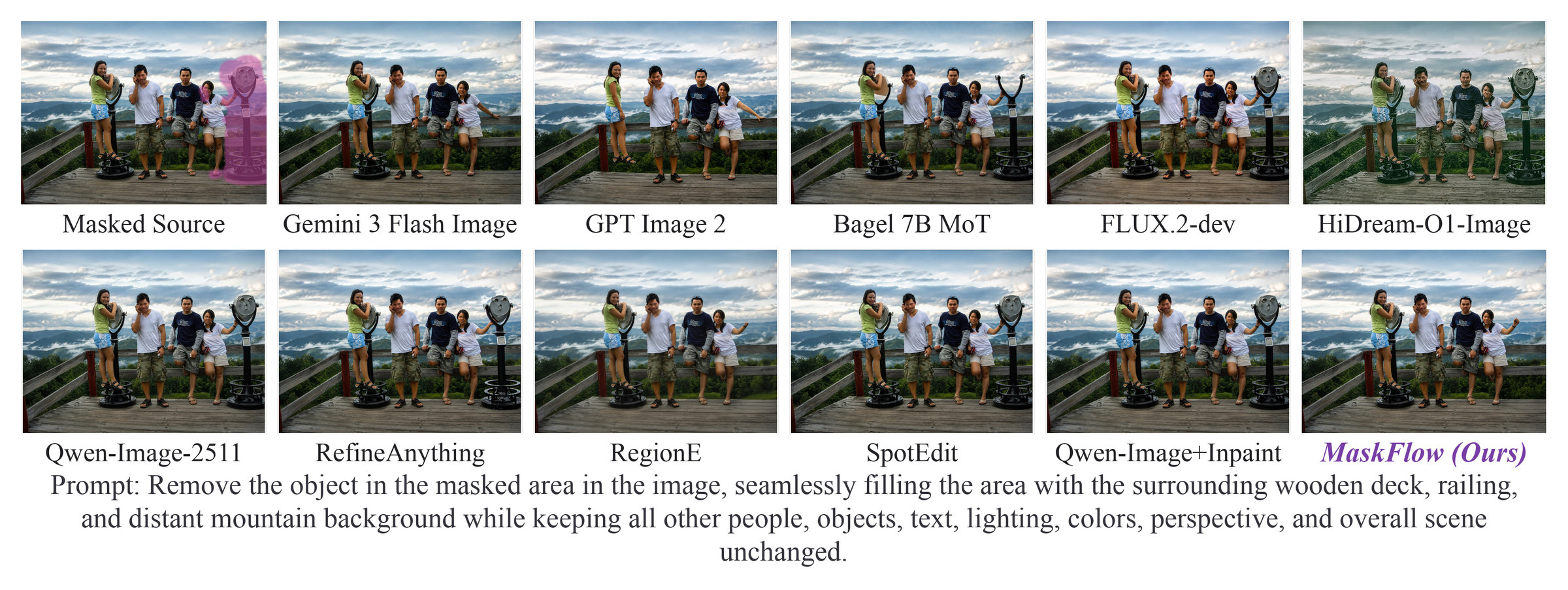}%
      }\par\vspace{2pt}
      \subfloat{%
        \includegraphics[width=\linewidth]{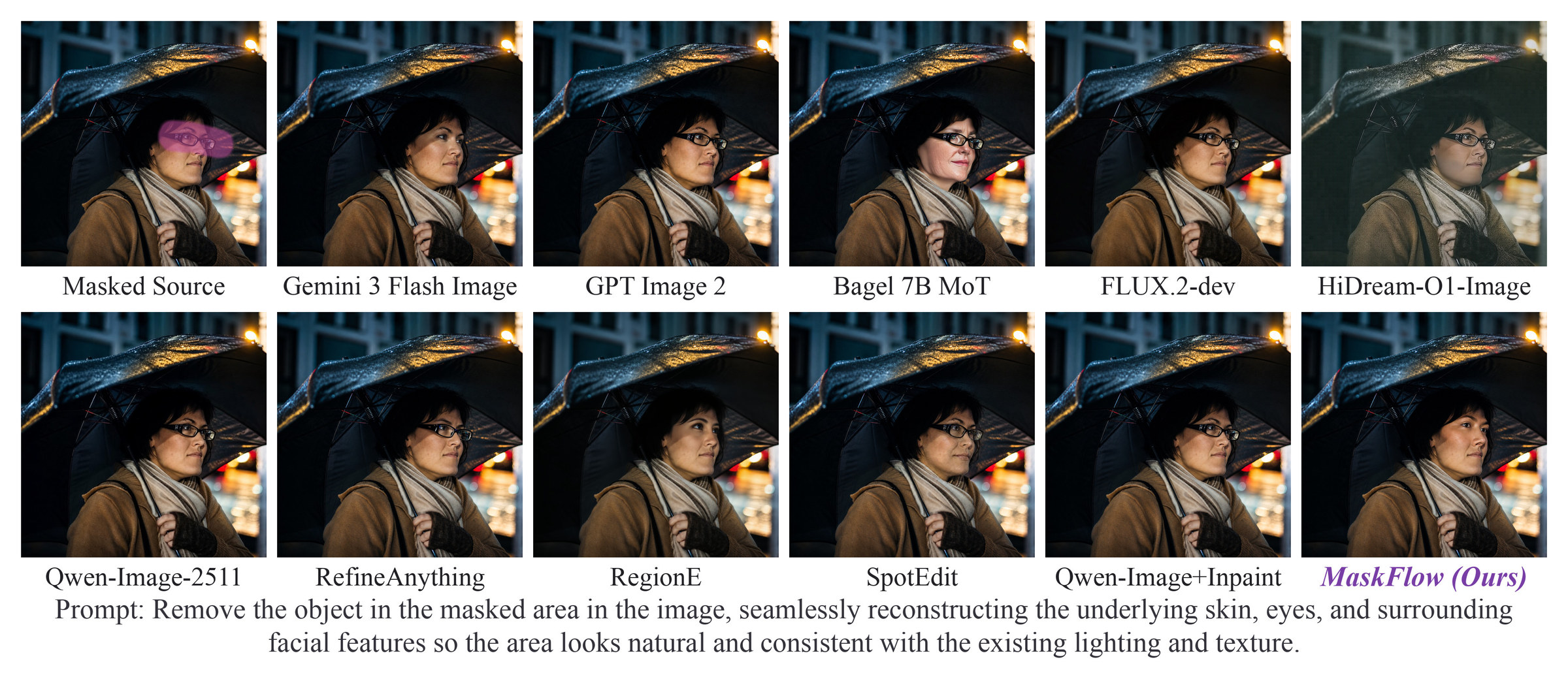}%
      }\par\vspace{2pt}
      \subfloat{%
        \includegraphics[width=\linewidth]{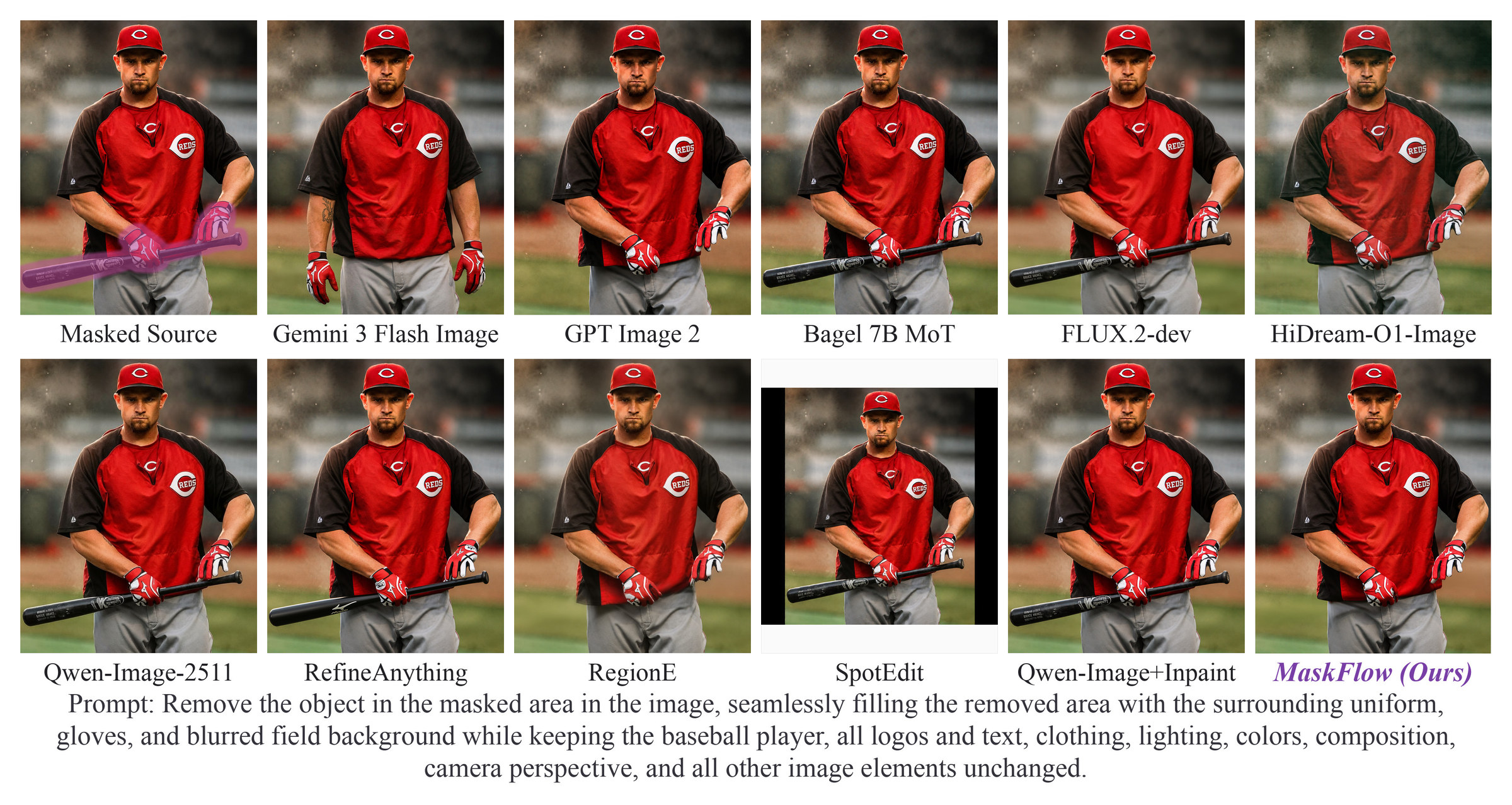}%
      }
    \end{minipage}
  \end{lrbox}
  \ifdim\dimexpr\ht\maskflowtaskbox+\dp\maskflowtaskbox\relax>\maskflowtaskheight
    \resizebox*{!}{\maskflowtaskheight}{\usebox{\maskflowtaskbox}}%
  \else
    \usebox{\maskflowtaskbox}%
  \fi
  \caption{Qualitative comparisons for removal on scene images.}
  \label{apx-fig:scene-remove}
\end{figure}

\clearpage

\subsection{Infographic Image Editing}
\label{apx:visualization-infographics-en}

\subsubsection{Addition}

\setlength{\maskflowtaskheight}{\dimexpr\textheight-\pagetotal-55pt\relax}
\begin{figure}[!ht]
  \centering
  \begin{lrbox}{\maskflowtaskbox}
    \begin{minipage}{\linewidth}
      \centering
      \subfloat{%
        \includegraphics[width=\linewidth]{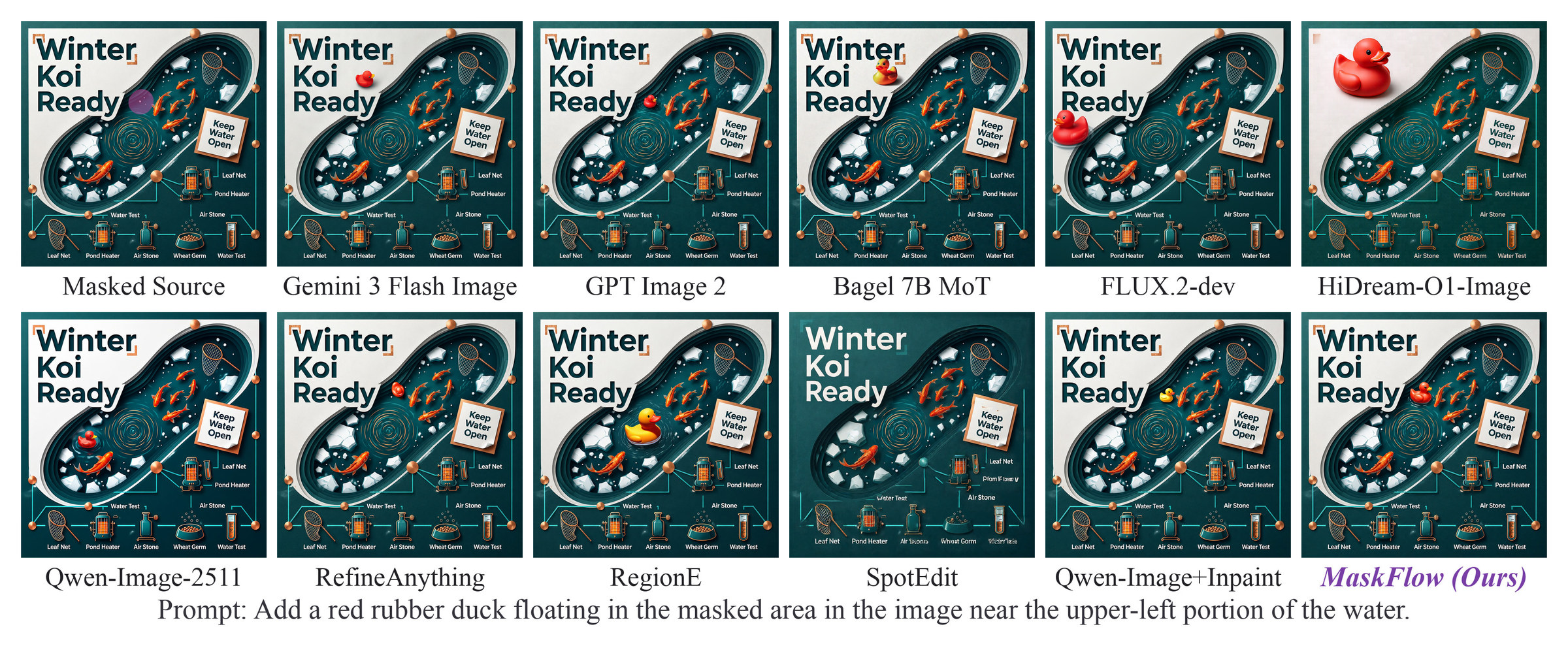}%
      }\par\vspace{2pt}
      \subfloat{%
        \includegraphics[width=\linewidth]{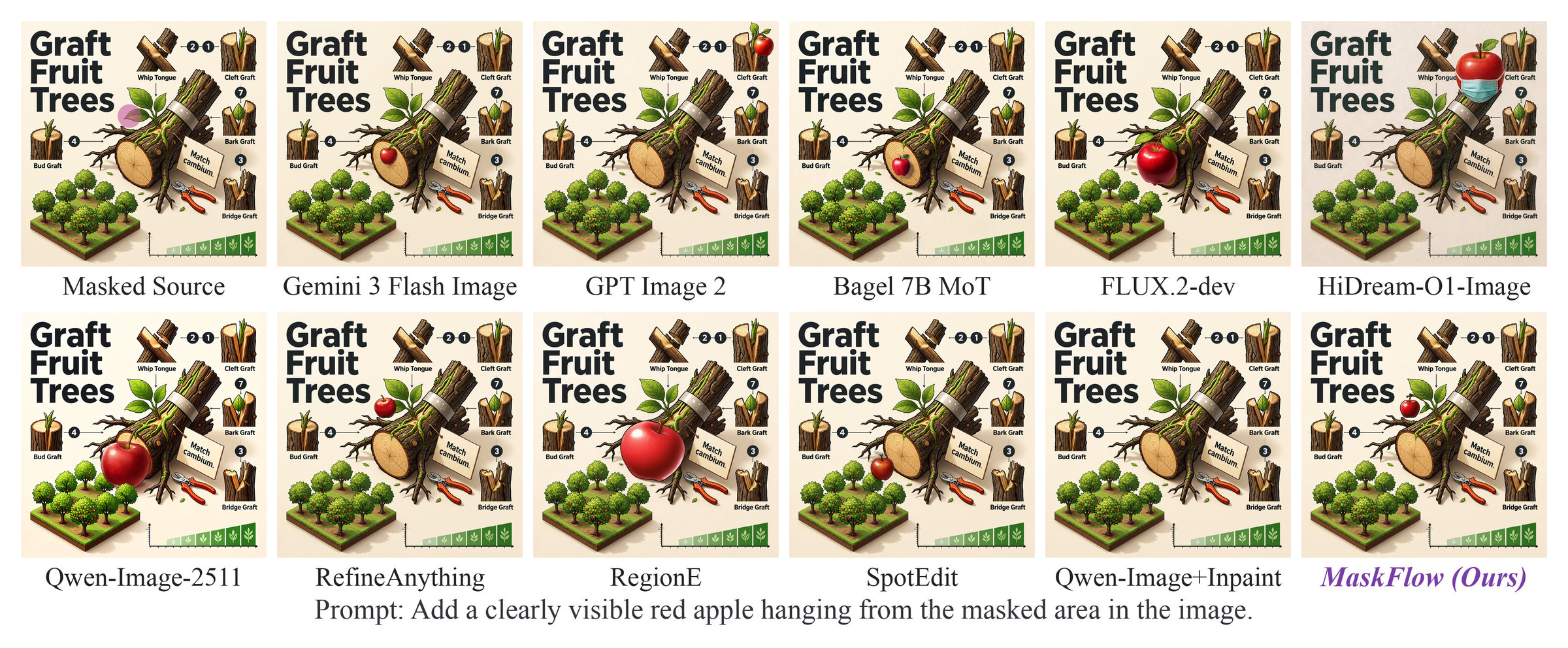}%
      }\par\vspace{2pt}
      \subfloat{%
        \includegraphics[width=\linewidth]{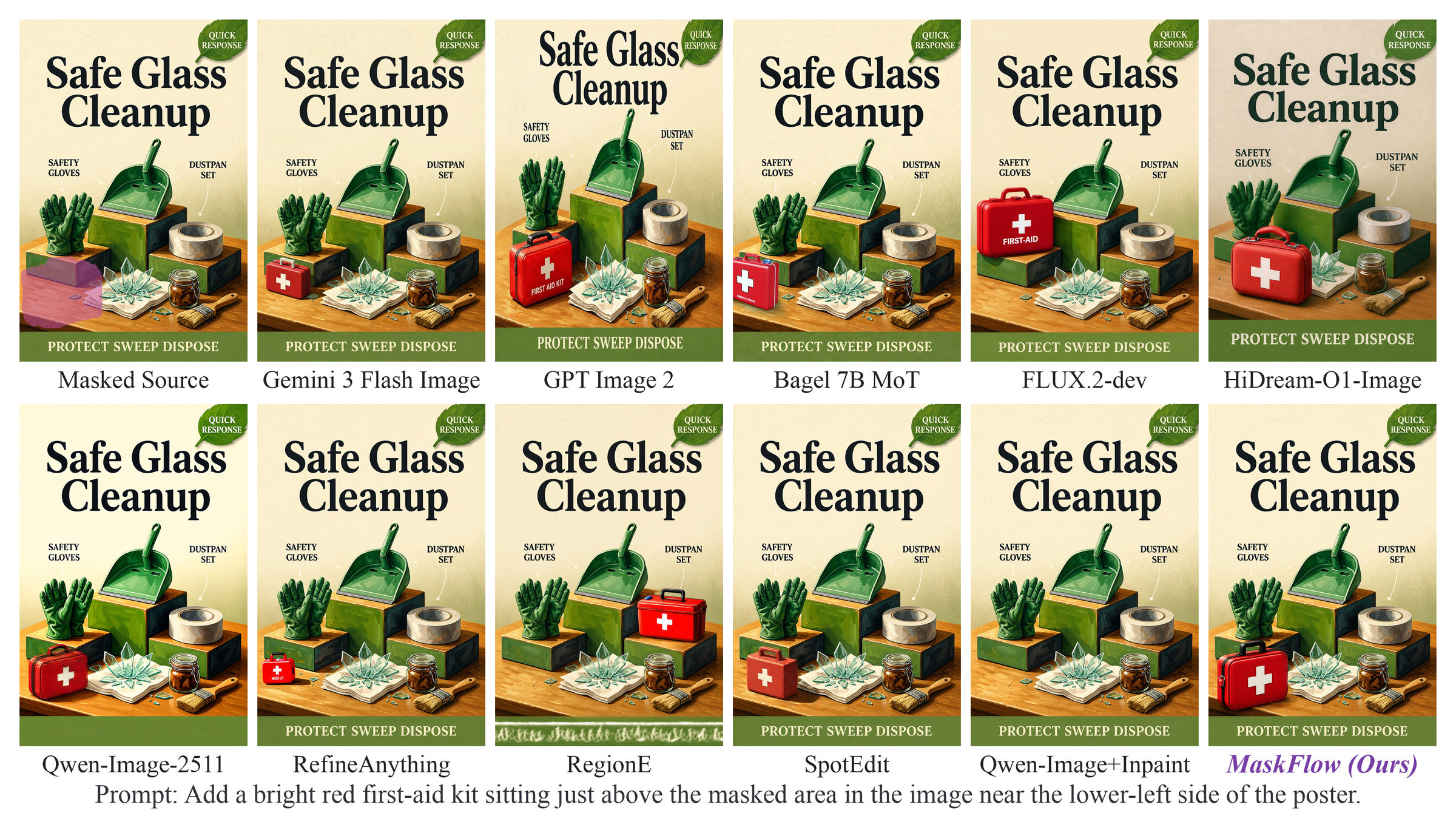}%
      }\par\vspace{2pt}
      \subfloat{%
        \includegraphics[width=\linewidth]{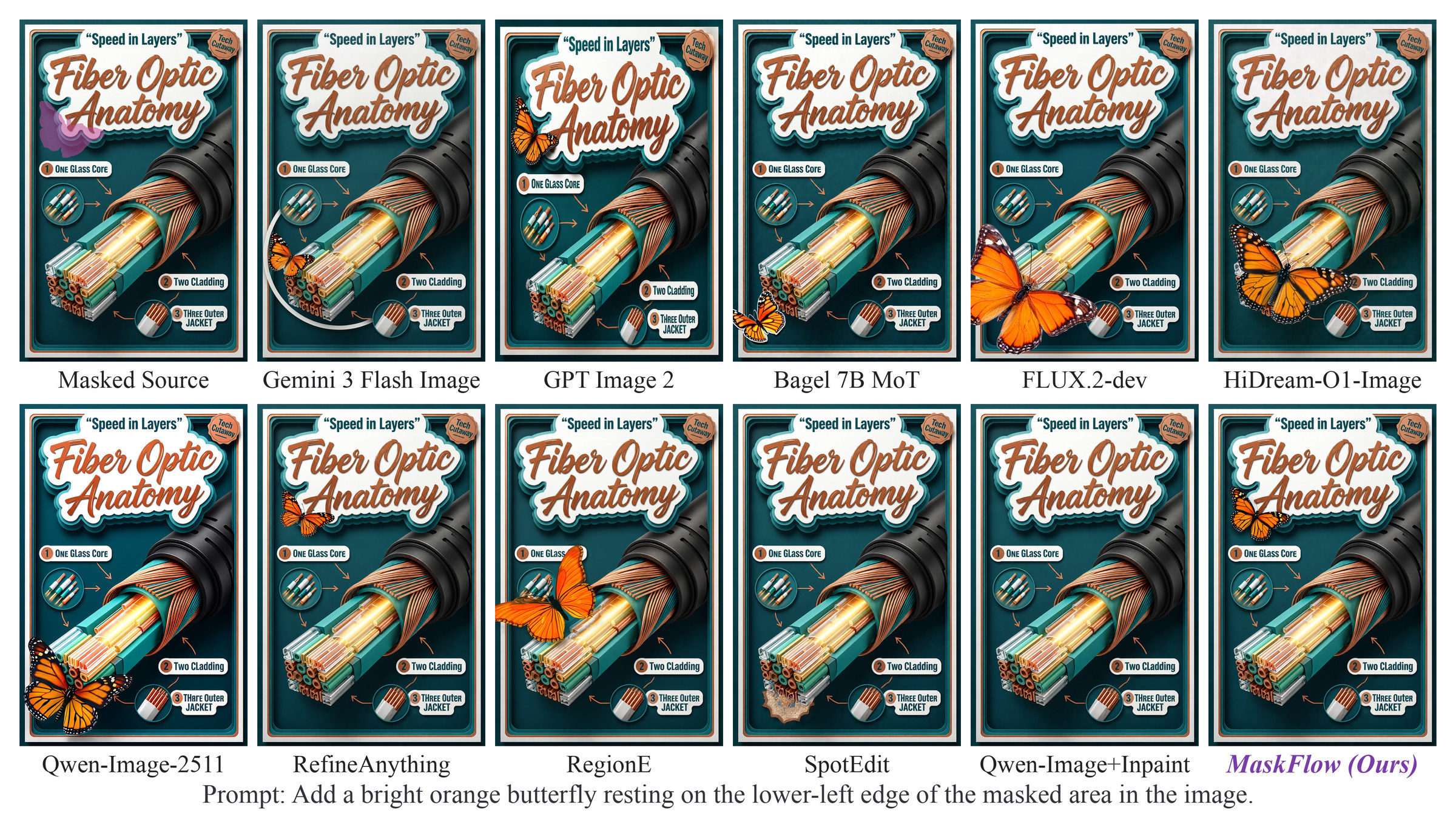}%
      }
    \end{minipage}
  \end{lrbox}
  \ifdim\dimexpr\ht\maskflowtaskbox+\dp\maskflowtaskbox\relax>\maskflowtaskheight
    \resizebox*{!}{\maskflowtaskheight}{\usebox{\maskflowtaskbox}}%
  \else
    \usebox{\maskflowtaskbox}%
  \fi
  \caption{Qualitative comparisons for addition on infographic images.}
  \label{apx-fig:infographics-en-add}
\end{figure}

\clearpage

\subsubsection{Modification}

\setlength{\maskflowtaskheight}{\dimexpr\textheight-\pagetotal-55pt\relax}
\begin{figure}[!ht]
  \centering
  \begin{lrbox}{\maskflowtaskbox}
    \begin{minipage}{\linewidth}
      \centering
      \subfloat{%
        \includegraphics[width=\linewidth]{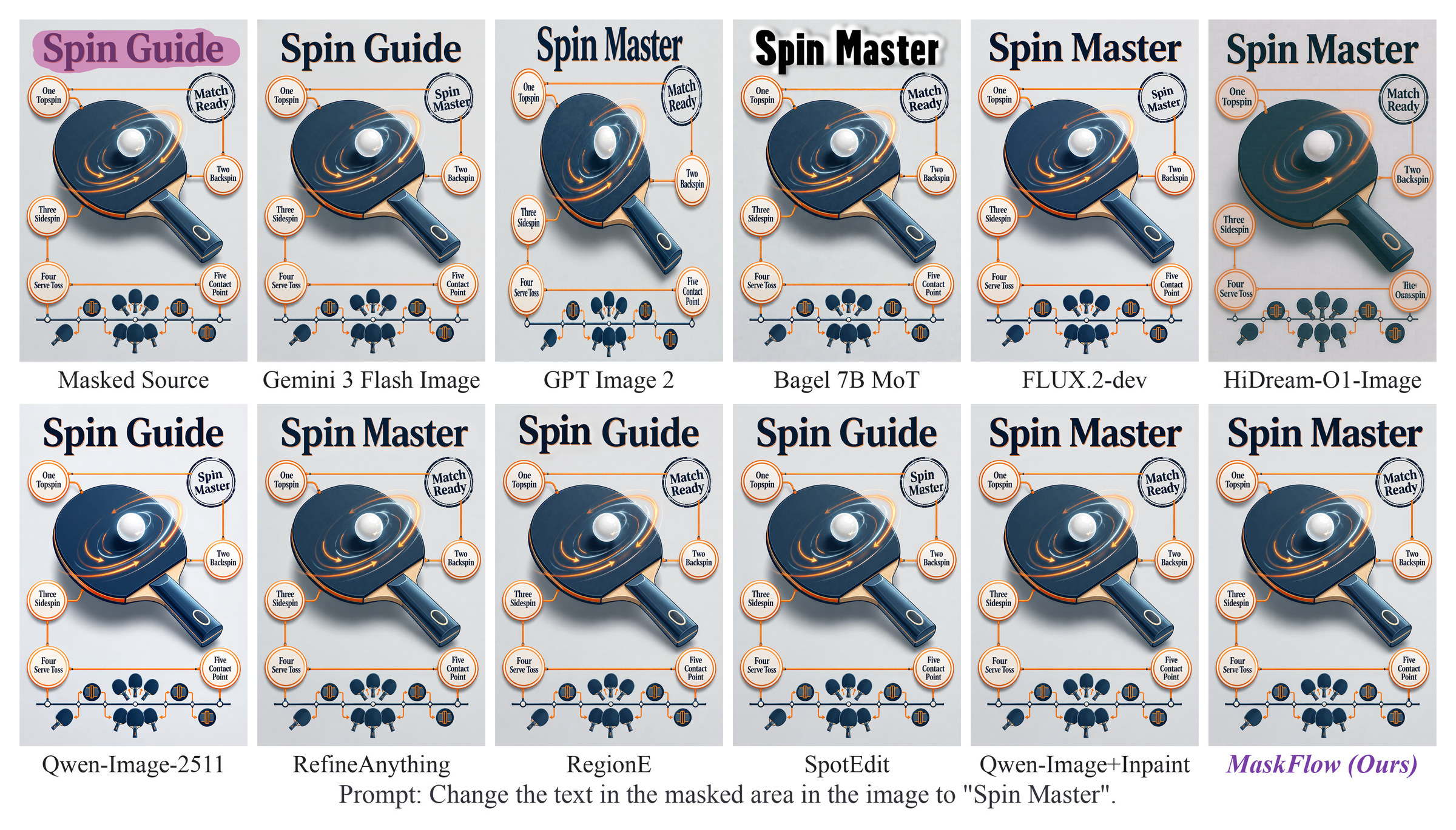}%
      }\par\vspace{2pt}
      \subfloat{%
        \includegraphics[width=\linewidth]{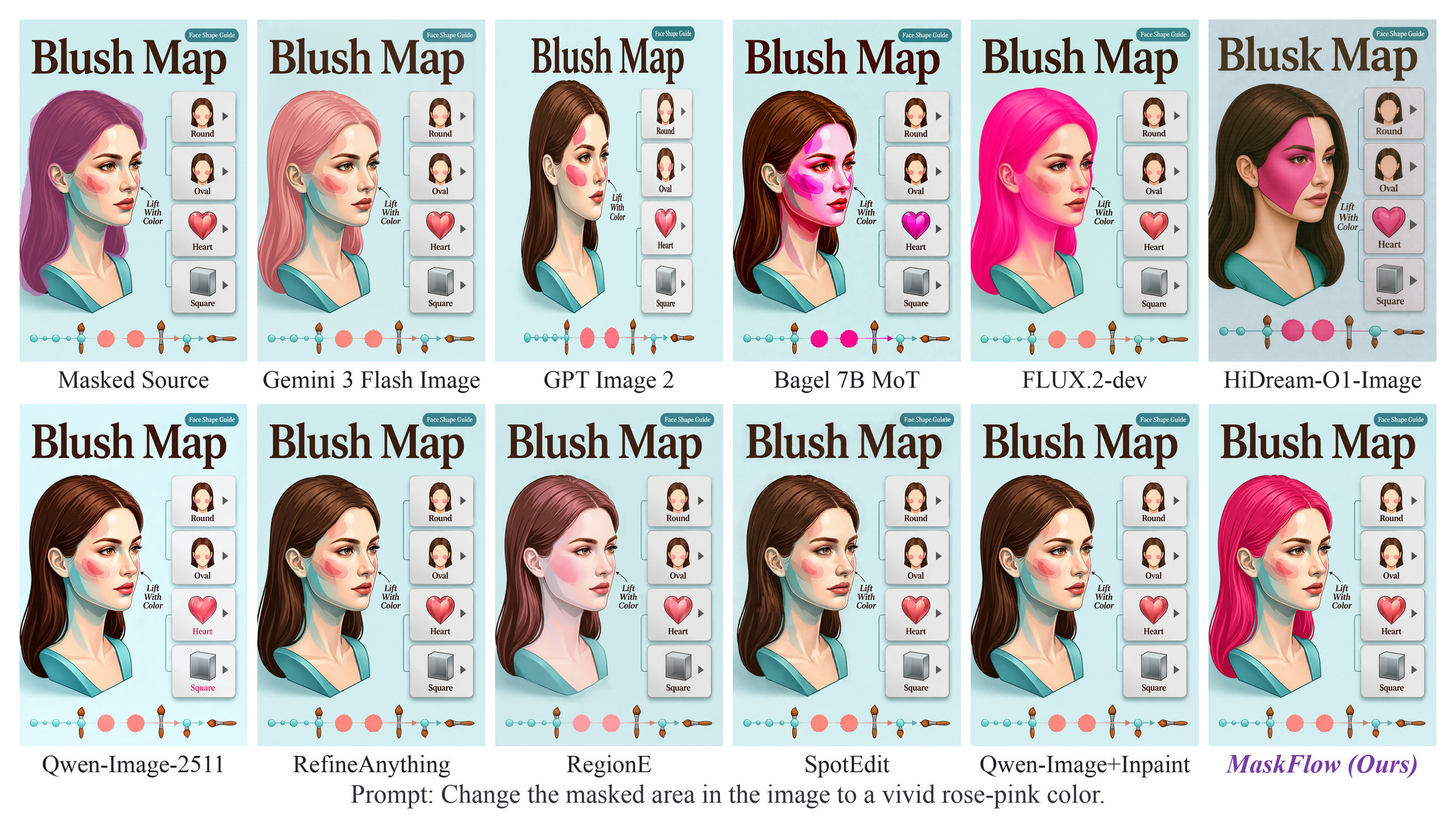}%
      }\par\vspace{2pt}
      \subfloat{%
        \includegraphics[width=\linewidth]{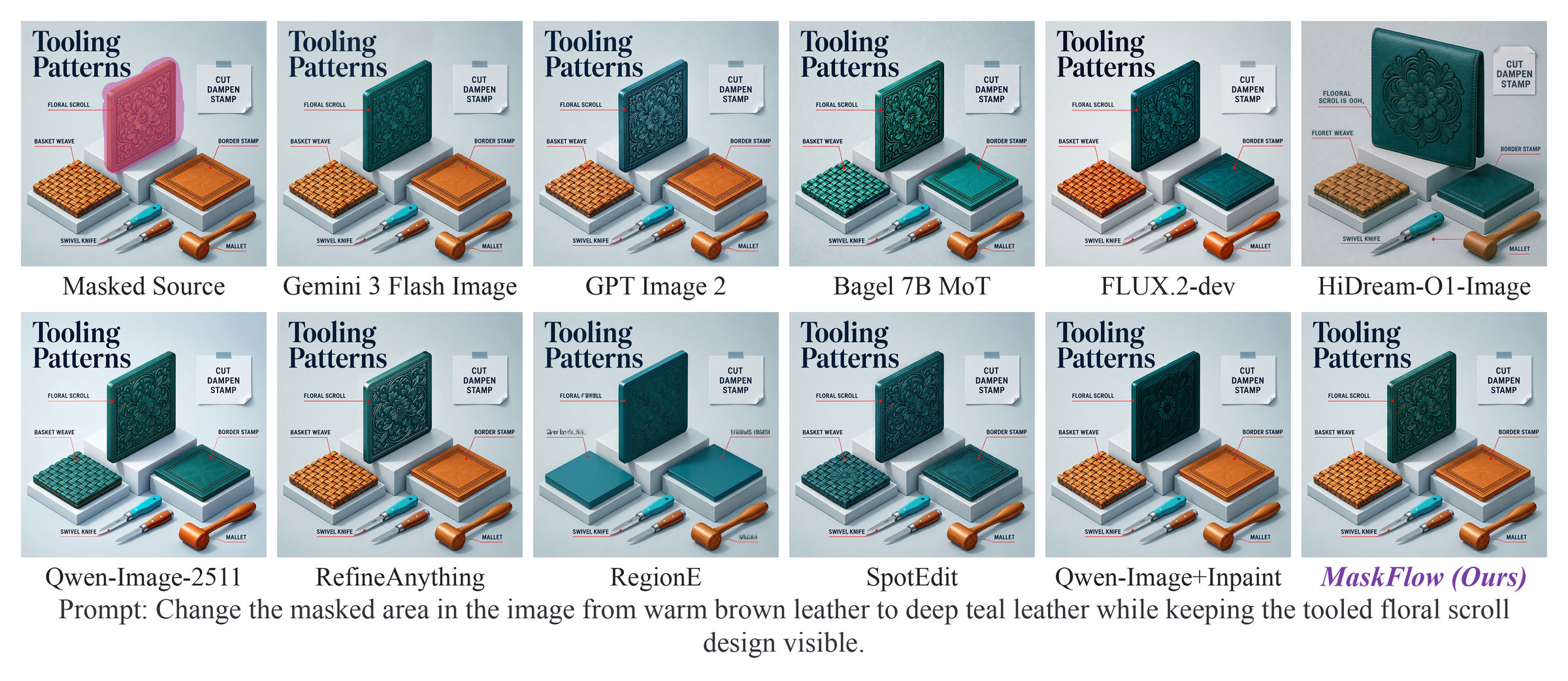}%
      }\par\vspace{2pt}
      \subfloat{%
        \includegraphics[width=\linewidth]{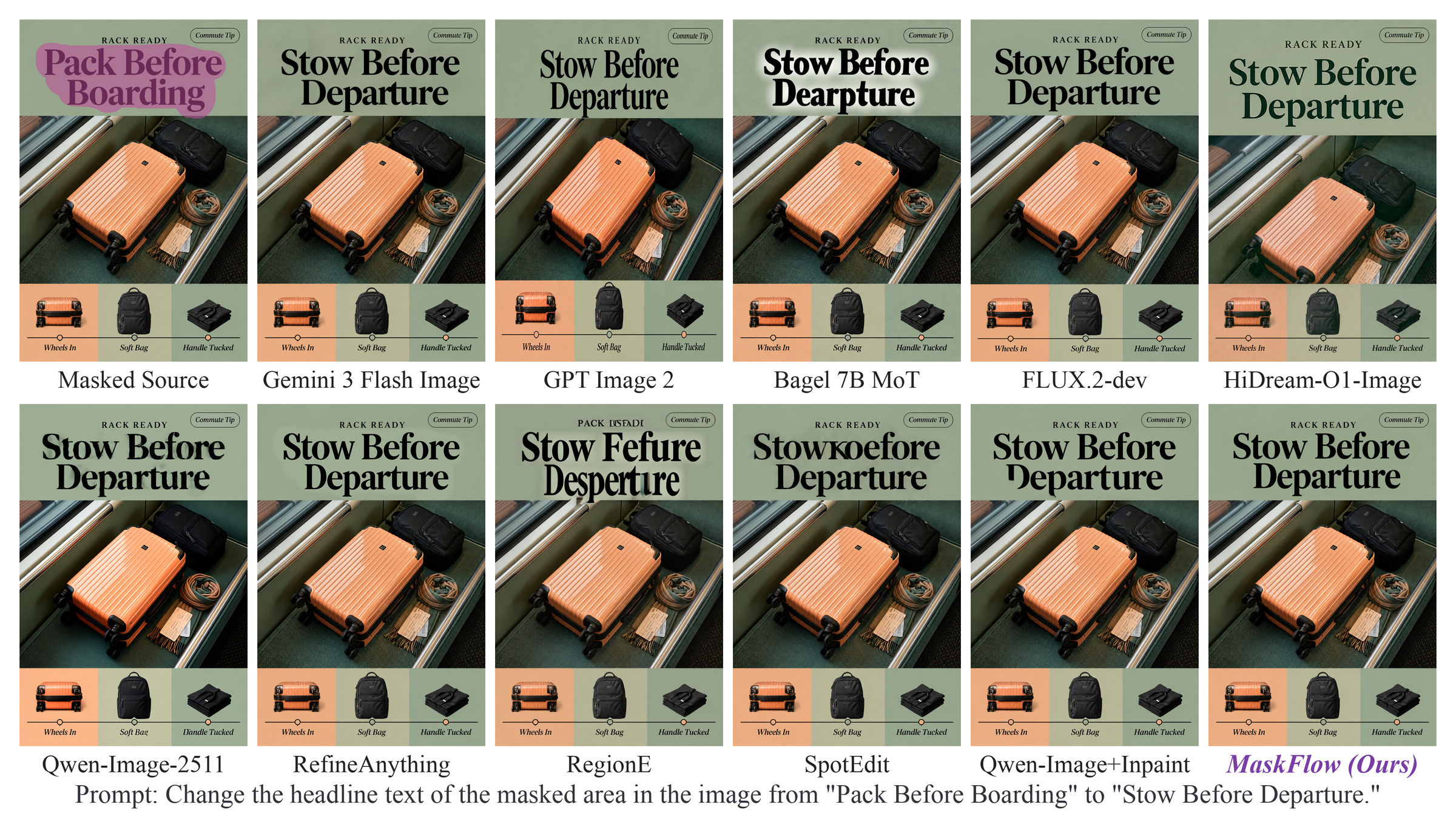}%
      }
    \end{minipage}
  \end{lrbox}
  \ifdim\dimexpr\ht\maskflowtaskbox+\dp\maskflowtaskbox\relax>\maskflowtaskheight
    \resizebox*{!}{\maskflowtaskheight}{\usebox{\maskflowtaskbox}}%
  \else
    \usebox{\maskflowtaskbox}%
  \fi
  \caption{Qualitative comparisons for modification on infographic images.}
  \label{apx-fig:infographics-en-modify}
\end{figure}

\clearpage

\subsubsection{Replacement}

\setlength{\maskflowtaskheight}{\dimexpr\textheight-\pagetotal-55pt\relax}
\begin{figure}[!ht]
  \centering
  \begin{lrbox}{\maskflowtaskbox}
    \begin{minipage}{\linewidth}
      \centering
      \subfloat{%
        \includegraphics[width=\linewidth]{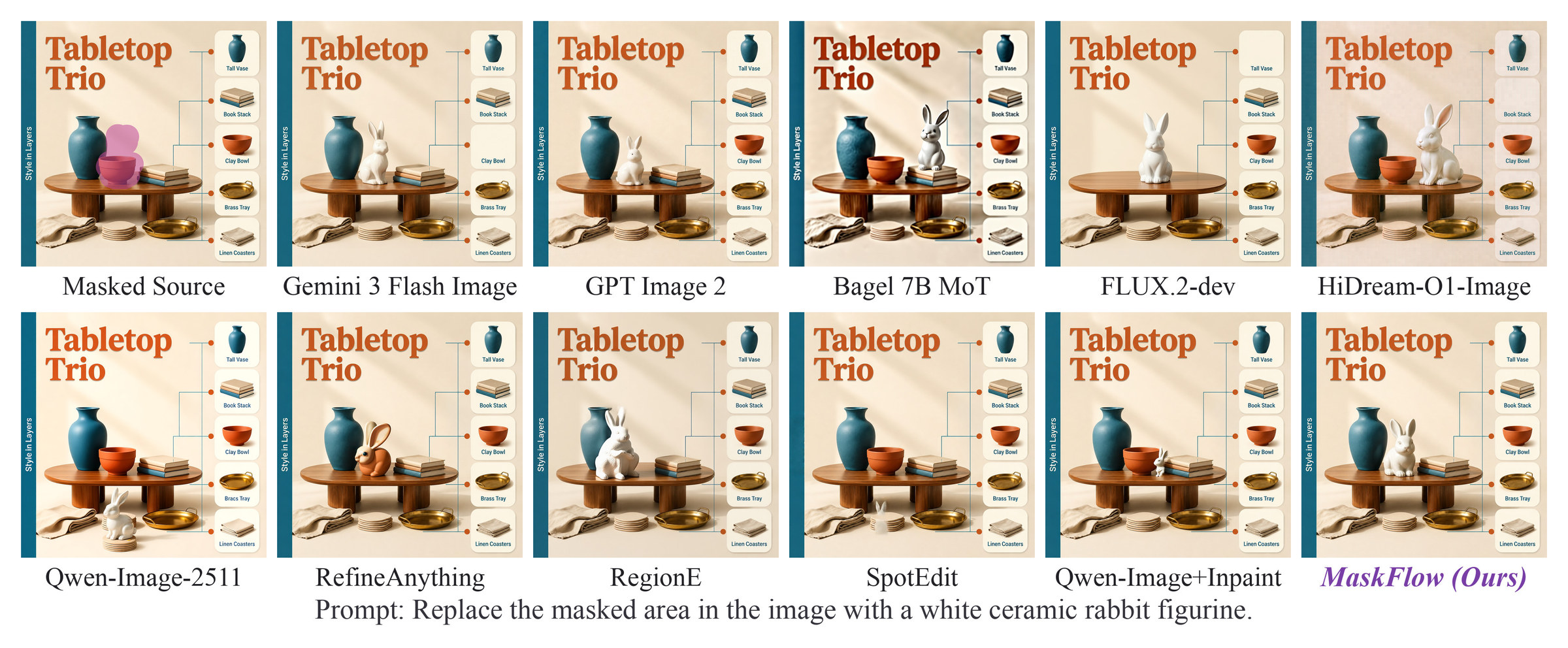}%
      }\par\vspace{2pt}
      \subfloat{%
        \includegraphics[width=\linewidth]{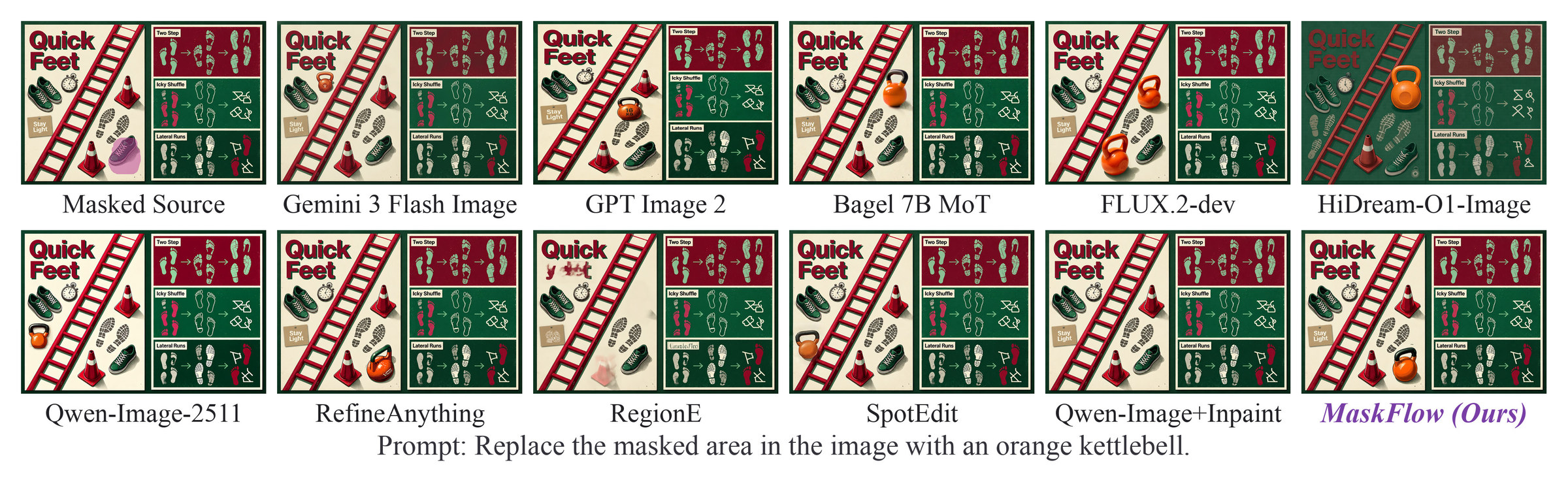}%
      }\par\vspace{2pt}
      \subfloat{%
        \includegraphics[width=\linewidth]{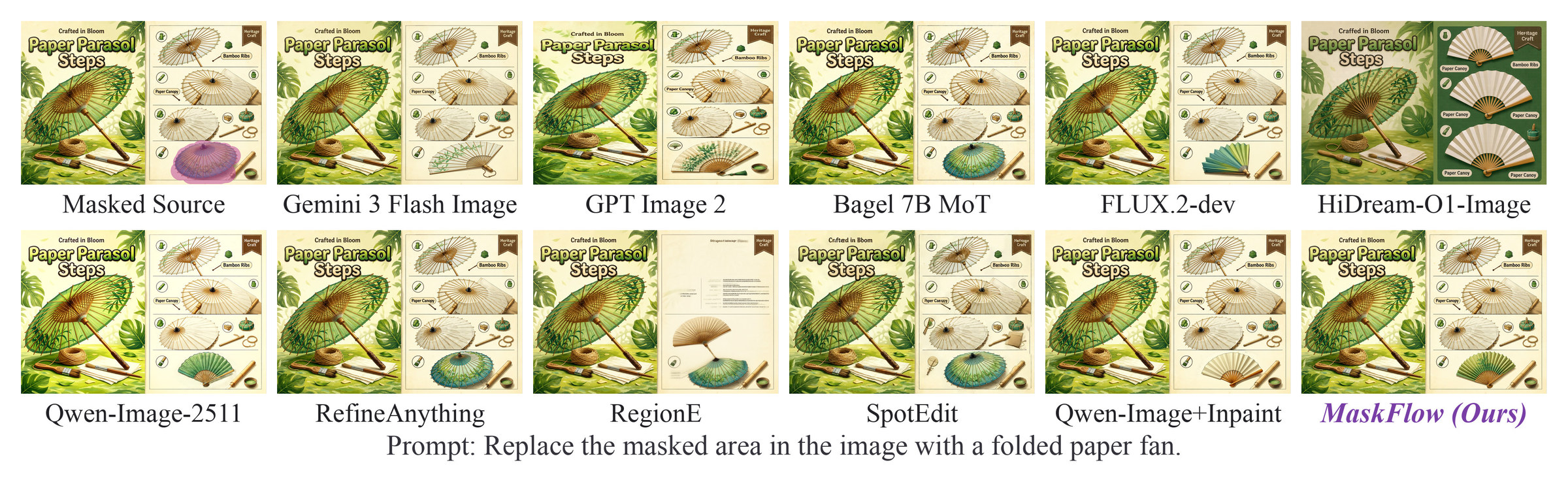}%
      }\par\vspace{2pt}
      \subfloat{%
        \includegraphics[width=\linewidth]{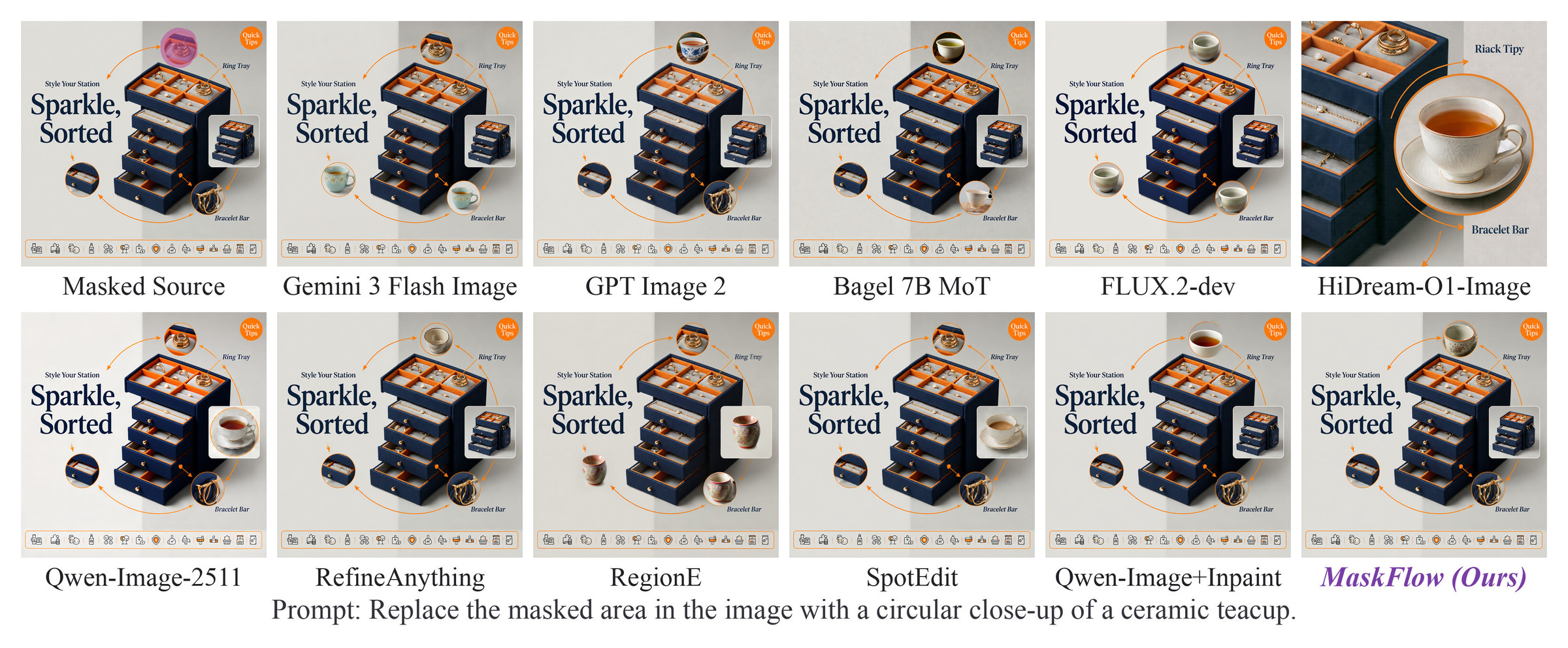}%
      }
    \end{minipage}
  \end{lrbox}
  \ifdim\dimexpr\ht\maskflowtaskbox+\dp\maskflowtaskbox\relax>\maskflowtaskheight
    \resizebox*{!}{\maskflowtaskheight}{\usebox{\maskflowtaskbox}}%
  \else
    \usebox{\maskflowtaskbox}%
  \fi
  \caption{Qualitative comparisons for replacement on infographic images.}
  \label{apx-fig:infographics-en-replace}
\end{figure}

\clearpage

\subsubsection{Removal}

\setlength{\maskflowtaskheight}{\dimexpr\textheight-\pagetotal-55pt\relax}
\begin{figure}[!ht]
  \centering
  \begin{lrbox}{\maskflowtaskbox}
    \begin{minipage}{\linewidth}
      \centering
      \subfloat{%
        \includegraphics[width=\linewidth]{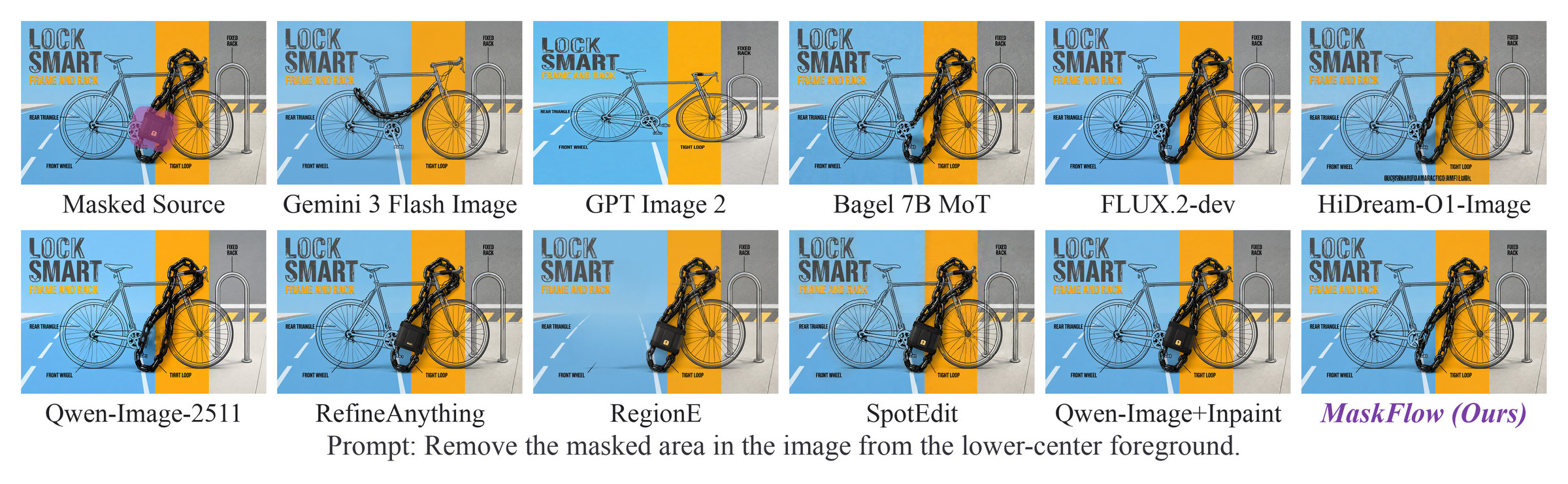}%
      }\par\vspace{2pt}
      \subfloat{%
        \includegraphics[width=\linewidth]{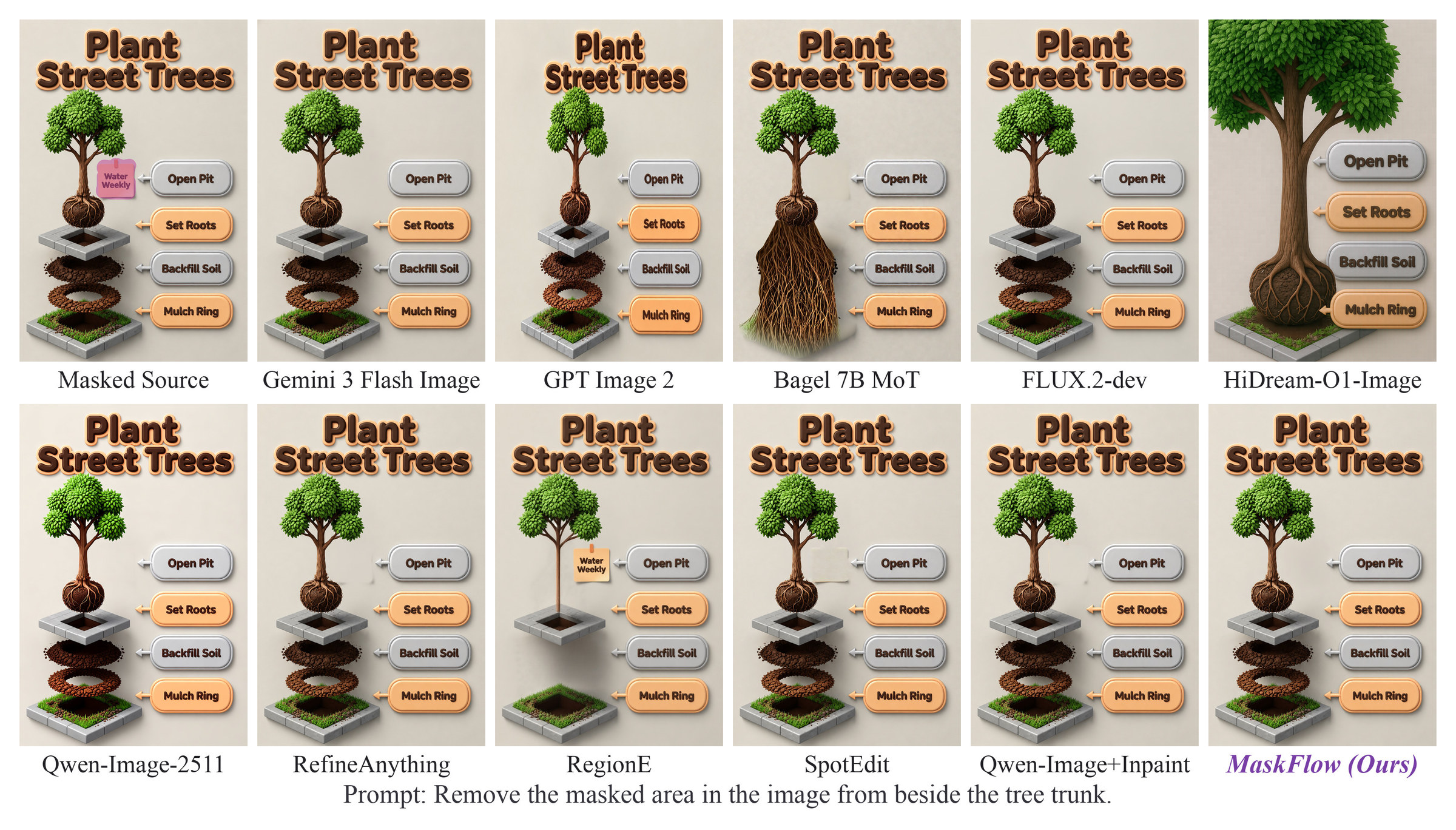}%
      }\par\vspace{2pt}
      \subfloat{%
        \includegraphics[width=\linewidth]{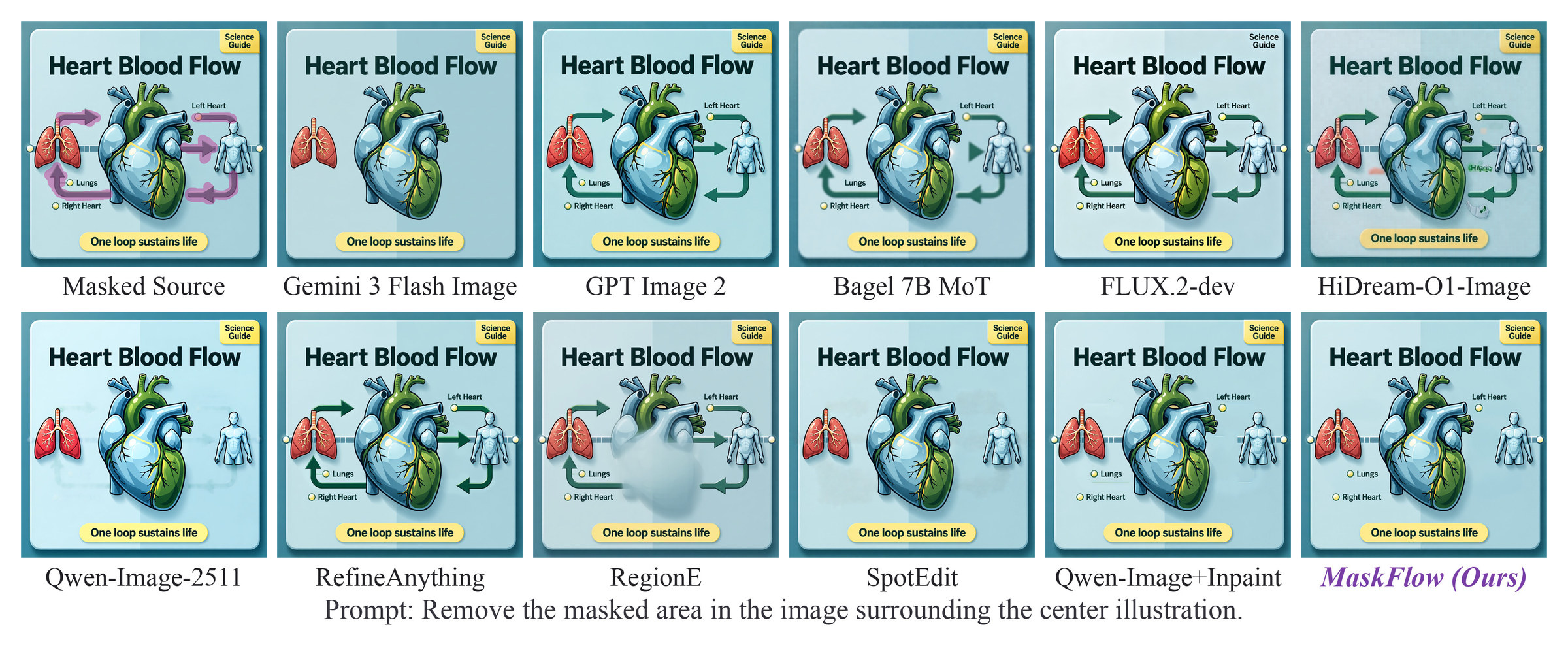}%
      }\par\vspace{2pt}
      \subfloat{%
        \includegraphics[width=\linewidth]{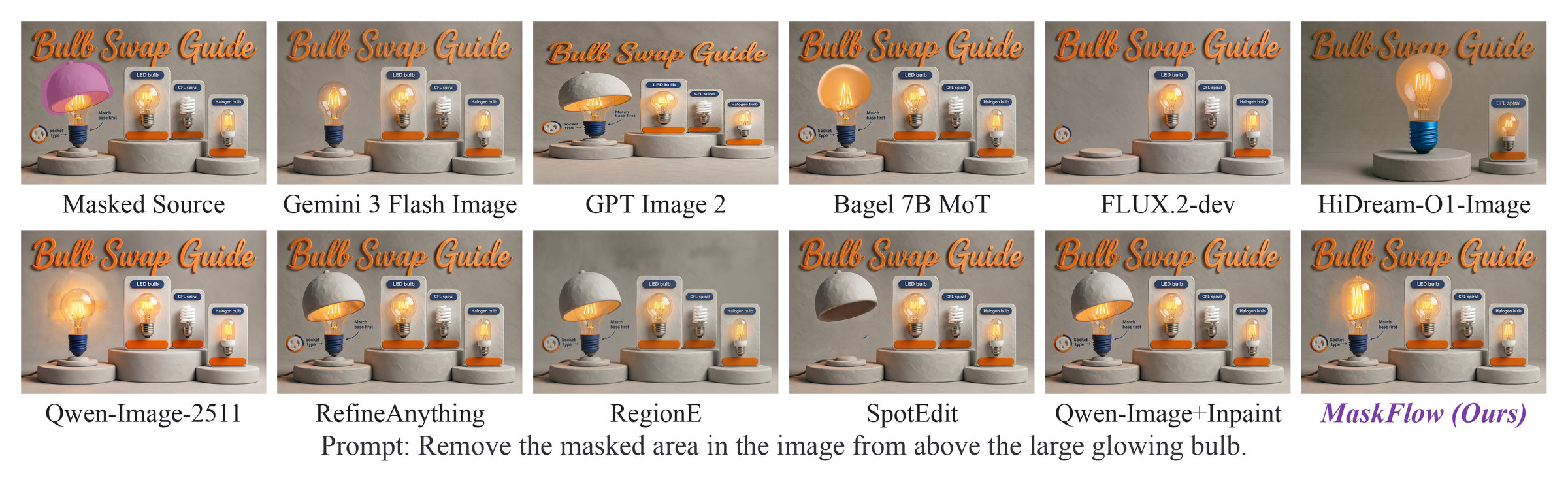}%
      }
    \end{minipage}
  \end{lrbox}
  \ifdim\dimexpr\ht\maskflowtaskbox+\dp\maskflowtaskbox\relax>\maskflowtaskheight
    \resizebox*{!}{\maskflowtaskheight}{\usebox{\maskflowtaskbox}}%
  \else
    \usebox{\maskflowtaskbox}%
  \fi
  \caption{Qualitative comparisons for removal on infographic images.}
  \label{apx-fig:infographics-en-remove}
\end{figure}

\end{document}